\documentclass[10pt,twocolumn,letterpaper]{article}

\usepackage{cvpr}              % To produce the CAMERA-READY version

\newcommand{\red}[1]{{\color{red}#1}}

\usepackage{amsmath} % preamble

\usepackage{enumitem}
\usepackage{multirow}
\usepackage{colortbl}
\usepackage{microtype}
\usepackage{comment}

\usepackage[hang,flushmargin]{footmisc}

\usepackage{amsfonts}

\usepackage{algorithm}
\usepackage{algpseudocode}
\usepackage{makecell}
\setlist[itemize]{align=parleft,left=0pt,topsep=1mm,itemsep=0mm,parsep=1mm}

\definecolor{azure(colorwheel)}{rgb}{0.0, 0.5, 1.0}
\definecolor{R5}{rgb}{0.0, 0.7, 0.1}
\definecolor{sohwi}{rgb}{0.01176, 0.5490, 0.5490}
\definecolor{R123}{rgb}{0.36, 0.54, 0.66}
\definecolor{R1234}{rgb}{0.7, 0.75, 0.71}
\definecolor{applegreen}{rgb}{0.55, 0.71, 0.0}
\definecolor{R132}{rgb}{0.0, 0.0, 1.0}
\definecolor{postechred}{rgb}{0.784, 0.003, 0.313}
\definecolor{gu}{rgb}{0.5460, 0.1755, 0.2766}
\definecolor{el}{rgb}{0.9764, 0.447, 0.447}
\definecolor{hyos}{rgb}{0.662, 0.482, 0.960}
\definecolor{joon}{rgb}{1.0, 0.294, 0.200}
\definecolor{ballblue}{rgb}{0.13, 0.67, 0.8}
\definecolor{cornellred}{rgb}{0.7, 0.11, 0.11}
\definecolor{darkcyan}{rgb}{0.0, 0.55, 0.55}
\definecolor{CuGray}{gray}{0.9}
\definecolor{airforceblue}{rgb}{0.36, 0.54, 0.66}
\definecolor{rev}{rgb}{0.784, 0.003, 0.313}
\definecolor{pink}{cmyk}{0, 0.7808, 0.4429, 0.1412}
\definecolor{amethyst}{rgb}{0.6, 0.4, 0.8}
\definecolor{black}{rgb}{0.0, 0.0, 0.0}
\definecolor{tb3_yellow}{rgb}{0.996, 1.0, 0.6}
\definecolor{R123}{rgb}{0.980, 0.8, 0.604}
\definecolor{R512}{rgb}{0.972, 0.6, 0.6}
\definecolor{dimgray}{rgb}{0.41, 0.41, 0.41}
\definecolor{R3}{rgb}{0.8, 0.25, 0.33}
\definecolor{bleudefrance}{rgb}{0.19, 0.55, 0.91}
\definecolor{R6}{rgb}{0.265, 0.445, 0.765}
\definecolor{blue(ryb)}{rgb}{0.01, 0.28, 1.0}
\definecolor{R4}{rgb}{1.0, 0.49, 0.0}
\definecolor{Gray}{gray}{0.88}
\definecolor{green(ncs)}{rgb}{0.0, 0.62, 0.42}
\definecolor{brightpink}{rgb}{1.0, 0.0, 0.5}
\definecolor{alizarin}{rgb}{0.82, 0.1, 0.26}
\definecolor{coral}{rgb}{0.9,0.32,0.30}

\definecolor{softblue}{rgb}{0.75, 0.76, 1.0}
\definecolor{softblueLight}{rgb}{0.88, 0.89, 1.00}

\definecolor{iclr_red}{rgb}{0.945, 0.651, 0.659}
\definecolor{iclr_blue}{rgb}{0.4392,0.2902,0.8235}
\definecolor{bubbles}{rgb}{0.91, 1.0, 1.0} 
\definecolor{blizzardblue}{rgb}{0.7, 0.95, 0.96}
\definecolor{bleudefrance}{rgb}{0.19, 0.55, 0.91}

\definecolor{kellygreen}{rgb}{0.3, 0.73, 0.09}

\newcolumntype{g}{>{\columncolor{CuGray}}c}
\newcolumntype{z}{>{\columncolor{CuGray}}l}

\renewcommand{\paragraph}[1]{\vspace{1mm}\noindent\textbf{#1.}\,}

\usepackage{xspace}

\makeatletter
\def\@fnsymbol#1{\ensuremath{\ifcase#1\or *\or \dagger\or \ddagger\or
   \mathsection\or \mathparagraph\or \|\or **\or \dagger\dagger
   \or \ddagger\ddagger \else\@ctrerr\fi}}
\makeatother

\def\onedot{.\@\xspace}
\def\eg{\emph{e.g}\onedot} 
\def\ie{\emph{i.e}\onedot}

\newcommand{\be}{\begin{eqnarray}}
\newcommand{\ee}{\end{eqnarray}}
\newcommand{\bee}{\begin{eqnarray*}}
\newcommand{\eee}{\end{eqnarray*}}

\newcommand{\matrixb}{\left[ \begin{array}}
\newcommand{\matrixe}{\end{array} \right]}

\usepackage{amssymb}
\usepackage{pifont}
\newcommand{\cmark}{\ding{51}}%
\newcommand{\xmark}{\ding{55}}%

\newcommand{\stoptocwriting}{%
  \addtocontents{toc}{\protect\setcounter{tocdepth}{-5}}}
\newcommand{\resumetocwriting}{%
  \addtocontents{toc}{\protect\setcounter{tocdepth}{\arabic{tocdepth}}}}

\usepackage{caption}
\usepackage{tabularx}
\usepackage{adjustbox}

\definecolor{cvprblue}{rgb}{0.21,0.49,0.74}
\usepackage[pagebackref,breaklinks,colorlinks,allcolors=cvprblue]{hyperref}
\usepackage{tabularx}

\def\paperID{} % *** Enter the Paper ID here
\def\confName{CVPR}
\def\confYear{2026}

\title{\textcolor[HTML]{ff8b8b}{C}\textcolor[HTML]{f4ab44}{L}\textcolor[HTML]{74d674}{A}\textcolor[HTML]{9195fa}{Y}: \textcolor[HTML]{ff8b8b}{C}onditiona\textcolor[HTML]{f4ab44}{l} Visual Simil\textcolor[HTML]{74d674}{a}rit\textcolor[HTML]{9195fa}{y} Modulation\\in Vision-Language Embedding Space}

\author{
Sohwi Lim \quad
Lee Hyoseok \quad
Jungjoon Park \quad
Tae-Hyun Oh \\[1em]
{
KAIST
}
}

\begin{document}

\twocolumn[
{
\renewcommand\twocolumn[1][]{#1}
\maketitle
\vspace{-7mm}
\begin{center}
\centering
\captionsetup{type=figure}
\includegraphics[width=0.99\linewidth]{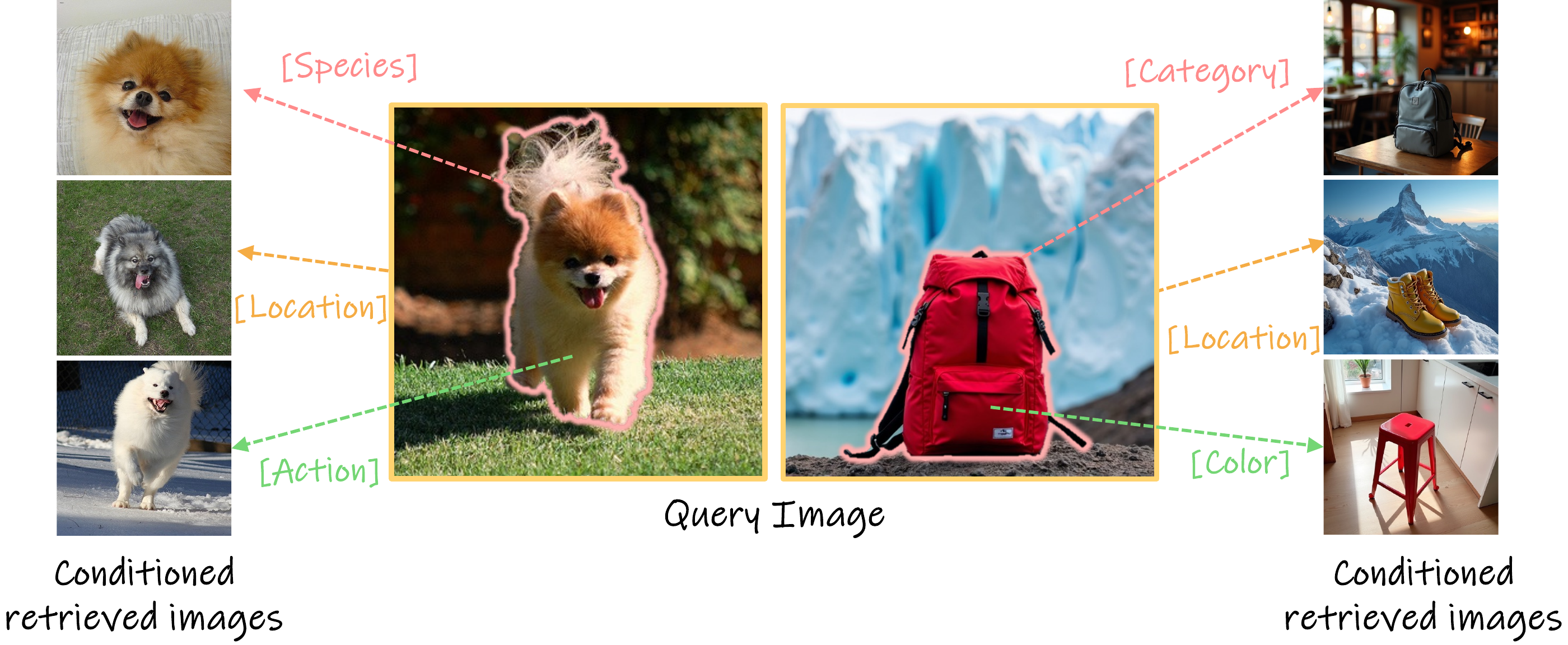}
    \vspace{-1mm}
   \captionof{figure}{Our proposed concept-based conditional image retrieval method retrieves images focusing on the semantic aspects specified by the text condition.
   Given a query image, our method adaptively computes conditioned similarity by modulating the similarity space to align with various conditions,~\eg, \texttt{species}, \texttt{location}, \texttt{action}, \texttt{category}, and \texttt{color}.
} 
\label{fig:teaser}
\vspace{3mm}
\end{center}
}]

\stoptocwriting 
\maketitle
\begin{abstract}
Human perception of visual similarity is inherently adaptive and subjective, depending on the users’ interests and focus.
However, most image retrieval systems fail to reflect this flexibility, relying on a fixed, monolithic metric that cannot incorporate multiple conditions simultaneously.
To address this, we propose \textbf{CLAY}, an adaptive similarity computation method that reframes the embedding space of pretrained Vision-Language Models (VLMs) as a text-conditional similarity space without additional training.
This design separates the textual conditioning process and visual feature extraction, allowing highly efficient and multi-conditioned retrieval with fixed visual embeddings.
We also construct a synthetic evaluation dataset \textbf{CLAY-EVAL}, for comprehensive assessment under diverse conditioned retrieval settings.
Experiments on standard datasets and our proposed dataset show that CLAY achieves high retrieval accuracy and notable computational efficiency compared to previous works. 
\end{abstract}    
\section{Introduction}
\label{sec:intro}
In the era of unprecedented data scale, humans seek to efficiently and accurately identify the information of interest in the overwhelming data flow.
Within this context, retrieval serves as a computational mechanism that enables us to find what truly matters and what we need~\cite{bang2026beyond, kilrain2025finer, ryan2025improving}.
In the computer vision area, despite the remarkable advances in large-scale image retrieval tasks, most approaches still rely on static definitions of visual similarity~\cite{fu2023dreamsim, radford2021learning, caron2021emerging, zhai2023sigmoid}.
By contrast, humans perceive visual similarity in a flexible and adaptive manner, selectively focusing on different aspects of an image depending on a user's interest as in Fig.~\ref{fig:teaser}.
For example, one may seek the same object itself, while another may want to see the overall mood of the image.
This underscores the need for conditional retrieval systems that can reflect various human attention-related visual cues.

\begin{table}[t]
\caption{
Our contextual conditional similarity computation method provides high retrieval accuracy while maintaining efficiency under diverse conditions. We further support a multi-conditioned retrieval scheme, whereas previous works~\cite{vaze2023genecis, hsieh2025focallens} do not. $^{\mathbf{\dagger}}$ denotes a modified version of the GeneCIS, where database features are additionally incorporated with conditional text.
}\label{tab:teaser}
\vspace{-0.5em}
\centering
\renewcommand{\arraystretch}{1.2} 
    \resizebox{1\linewidth}{!}{
    \begin{tabular}{m{2.7cm} cccc}
    \toprule
     &  GeneCIS & GeneCIS$\mathbf{\dagger}$ & FocalLens & \textbf{\textcolor[HTML]{ff8b8b}{C}\textcolor[HTML]{f4ab44}{L}\textcolor[HTML]{74d674}{A}\textcolor[HTML]{9195fa}{Y}}\\
    \midrule
    Training-free &  \textcolor{red}{\xmark} &  \textcolor{red}{\xmark} & \textcolor{red}{\xmark} & \textcolor{green(ncs)}{\cmark}\\
    Retrieval accuracy & \textcolor{red}{\xmark}& \textcolor{green(ncs)}{\cmark}& \textcolor{green(ncs)}{\cmark}&\textcolor{green(ncs)}{\cmark}\\
    Dynamic efficiency &   \textcolor{green(ncs)}{\cmark}& \textcolor{red}{\xmark}& \textcolor{red}{\xmark}&\textcolor{green(ncs)}{\cmark}\\
    Multi-condition & \textcolor{red}{\xmark}& \textcolor{red}{\xmark} & \textcolor{red}{\xmark} & \textcolor{green(ncs)}{\cmark}\\
    \bottomrule
    \end{tabular}
    } 
\end{table}
In response to this demand, prior studies have explored two main directions depending on which aspects of the query image are focused or modified during retrieval: one focuses on particular attributes within the query image~\cite{hsieh2025focallens, vaze2023genecis, veit2017conditional}, 
while the other aims to change specific visual contexts of the query image to match the target images~\cite{baldrati2023zero, Saito_2023_CVPR,zhang2024magiclens, karthik2024vision, Gu_2024_CVPR, jang2024spherical}.
Although the latter direction has been actively studied, the former direction, which seeks to capture specific attributes in an adaptive manner, remains relatively underexplored despite its importance.
Existing studies to address it have primarily relied on training-based approaches, where models are trained to perform retrieval under given conditions by learning condition-specific representations~\cite{vaze2023genecis, hsieh2025focallens}.
However, these methods require not only computational resources for the training stage but also paired query-target image data for each condition~\cite{vaze2023genecis}, which limits their working regime to closed-set conditions.
Moreover, these trained conditional feature extractors incur computational overhead at inference time since all features of database images need to be recomputed through the model from scratch, whenever the user condition changes~\cite{veit2017conditional, hsieh2025focallens}.
These limitations hinder the applicability and capacity of conditional retrieval systems in diverse, real-world scenarios.

\begin{figure}[!t]
\centering
   \includegraphics[width=1\linewidth]{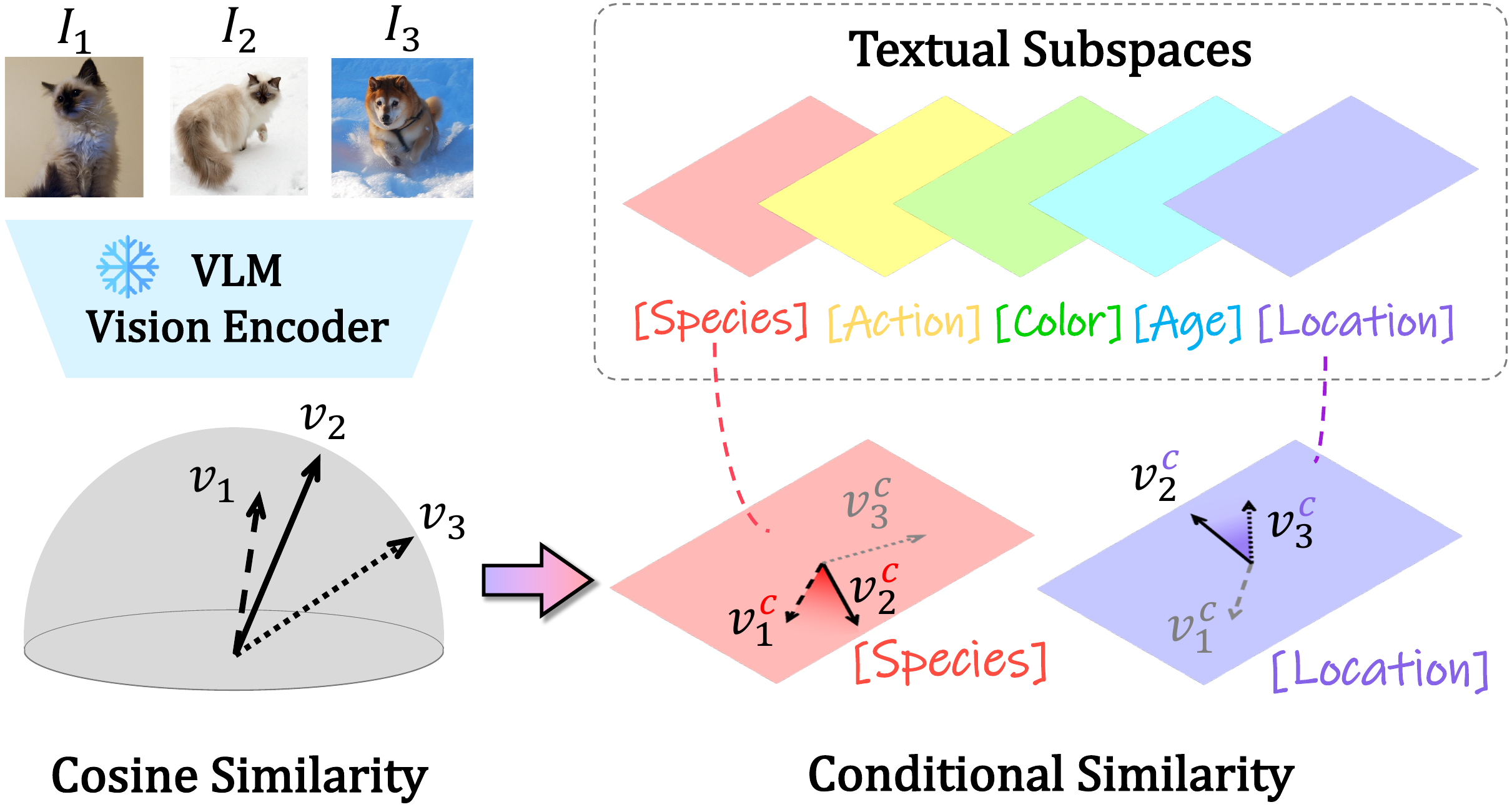}
   \vspace{-5mm}
    \caption{\textbf{Illustration of the concept of \textcolor[HTML]{ff8b8b}{C}\textcolor[HTML]{f4ab44}{L}\textcolor[HTML]{74d674}{A}\textcolor[HTML]{9195fa}{Y}.} Our method adaptively computes conditional similarity between images by modulating the original similarity space into a conditional similarity space within the representation space of VLMs.}
\label{fig:concept}
\end{figure}

To overcome these challenges, we propose \textcolor[HTML]{ff8b8b}{C}\textcolor[HTML]{f4ab44}{L}\textcolor[HTML]{74d674}{A}\textcolor[HTML]{9195fa}{Y}, a training-free and adaptive conditional similarity computation method that transforms the similarity computation space of pretrained Vision-Language Models~(VLMs)~\cite{radford2021learning, zhai2023sigmoid}, according to users' interest as in Fig.~\ref{fig:concept}.
By decoupling the conditioning procedure from visual feature extraction, our approach contextually modulates the similarity computation space of VLMs into a conditional similarity space.
This eliminates the need to completely recompute the embeddings of database images under varying user conditions by keeping the original visual embeddings fixed.
To achieve this, we construct a conditional similarity space within the VLMs’ representation space that respects its underlying non-Euclidean geometry, and define its textual concept subspace leveraging a subspace–construction procedure similar to~\cite{dorfman2025ip, nguyen2025csd}.
Building on this, we further demonstrate that our method is easily extensible to multi-conditional retrieval scenarios, unlike previous methods dedicated to single-condition settings~\cite{hsieh2025focallens, vaze2023genecis}.
Due to a lack of a standard benchmark of a multi-conditional retrieval task, we construct a synthetic evaluation dataset containing diverse human and object images and various conceptual condition pairs, allowing evaluations under multi-condition scenarios.
Evaluation on diverse real and synthetic datasets shows that our method achieves strong retrieval accuracy and notable computational efficiency, pushing the Pareto-front of the trade-off between performance and efficiency in practice.
Table~\ref{tab:teaser} summarizes the advantages of our method compared to previous approaches.
We summarize our contributions as:
\begin{itemize}
    \item We propose an efficient, training-free, state-of-the-art conditional visual similarity computation method that can contextually adapt to various conditions without recomputing database features at inference.
    \item Our method supports multi-conditioned image retrieval scenarios, enabling flexible retrieval settings beyond single-condition settings.
    \item We construct a synthetic evaluation dataset that contains diverse humans and objects with conceptual condition pairs to facilitate the evaluation under diverse conditional retrieval scenarios.
\end{itemize}

\section{Related Work}
\label{sec:related_work}
\noindent \textbf{Image-to-image retrieval} is a practical and fundamental problem in the computer vision area~\cite{gordo2016deep, hays2007scene, choi2026patch, liu2021cirr}.
Conventional methods on image-to-image retrieval mainly focused on measuring how visually similar two images are, relying on hand-crafted local descriptors~\cite{lowe2004distinctive,dalal2005histograms}.
With the emergence of deep learning, subsequent studies leveraged Convolutional Neural Networks~\cite{oquab2014learning, wei2016cross, Noh_2017_ICCV, cao2020unifying} to extract higher-level semantic features beyond low-level visual cues.
However, relying solely on static visual feature similarity from such models may be insufficient, as human preferences can vary across multiple contextual factors.
This limitation suggests that the retrieval systems need to be designed to consider further conditions, \ie, conditional image retrieval.
In this work, we explore conditional image retrieval, which considers adaptive alignment with user intentions.

\noindent \textbf{Vision-Language Embedding Models} are trained on large-scale image-text pair data to embed images and texts into a shared embedding space.
Vision-Language Models~(VLMs) such as CLIP~\cite{radford2021learning} or SigLIP~\cite{zhai2023sigmoid}, support semantic alignment of visual and textual representations through contrastive learning.
Multi-modal Large Language Models~(mLLMs) have extended to provide a more flexible representation conditioning process by utilizing the instruction-following capability of LLMs~\cite{meng2025vlm2vecv2, qwen3vlembedding, kim2026meol}.
Recent studies~\cite{dorfman2025ip, nguyen2025csd, Ye-Bin_2023_ICCV} have explored the structure of the VLM embedding space to enable text-guided control over visual embeddings for various tasks.
For instance, \citet{dorfman2025ip} project visual embedding into a text-based subspace, thereby extracting the text-related visual contexts useful for compositional image generation.
However, these approaches primarily focus on aligning visual representations with textual semantics, rather than modeling their relative relationships.
We further develop this idea to effectively capture the relationships among visual features within the textual subspace, accounting for the hyperspherical nature of the embedding manifold for accurate relationship modeling.

\begin{figure*}[!t]
\centering
   \includegraphics[width=1\linewidth]{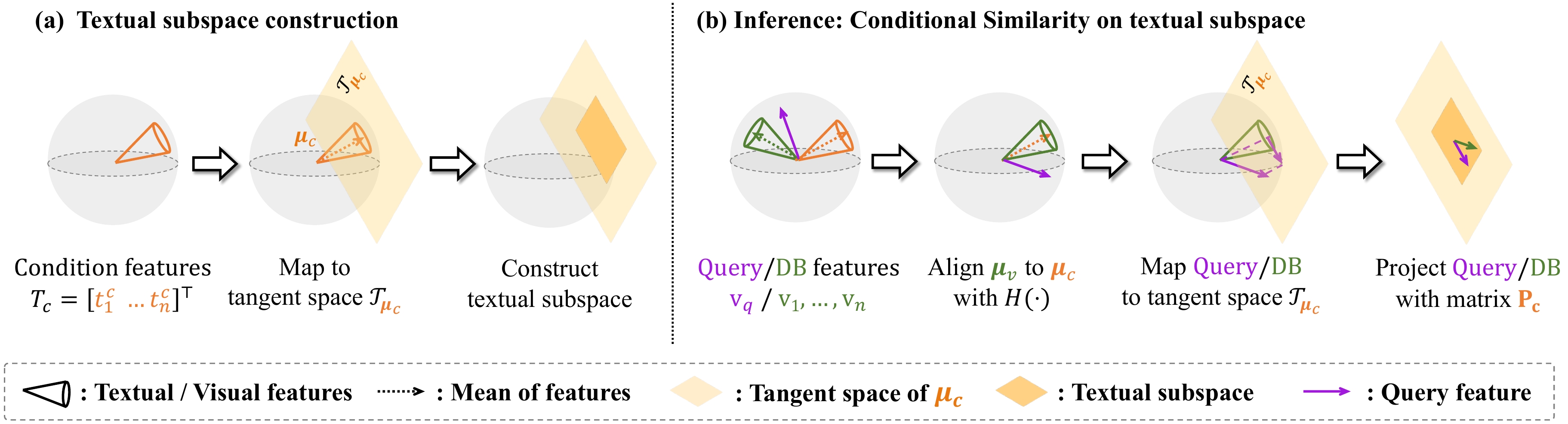}
    \caption{\textbf{Conditional similarity computation pipeline.} (a) Given a condition, we construct the manifold-aware textual subspace with the condition text features in advance, and generate the condition-aware projection matrix \textcolor[HTML]{edaf32}{$\mathbf{P}_c$}.
    (b) At inference, we compute the conditional similarity between the \textcolor[HTML]{a11adc}{query} and \textcolor[HTML]{548233}{database} images by projecting the visual features onto the textual subspace with \textcolor[HTML]{edaf32}{$\mathbf{P}_c$}.
    }
\label{fig:method}
\vspace{-3mm}
\end{figure*}
\paragraph{Conditional image retrieval}
With the increasing importance of retrieving images that satisfy user-specified conditions, prior studies have proposed utilizing text conditions and incorporating visual features to focus on specific attributes~\cite{hsieh2025focallens} in the image or to modify particular contexts~\cite{guo2019fashion, liu2021cirr, baldrati2023zero, zhang2024magiclens}.
As an early work, Conditional Similarity Networks~\cite{veit2017conditional} alleviated the limitation of the single similarity metric of embedding methods by employing conditioning masks to select condition-specific embedding dimensions.
To specify this task, GeneCIS~\cite{vaze2023genecis} introduced a benchmark that categorizes text conditions into two types: \textit{focus on} and \textit{change}, and proposed a training-based method that fine-tunes the image encoder and condition feature modulator from the paired dataset.
Following this, most subsequent studies have tackled the \textit{change} type, retrieving images that differ only in specific attributes from the query image, known as Composed Image Retrieval~(CIR)~\cite{guo2019fashion, liu2021cirr, baldrati2023zero, jang2024spherical, Gu_2024_CVPR, karthik2024vision, zhang2024magiclens}.
Another line of work~\cite {hsieh2025focallens} aims to retrieve images by focusing on specific attributes within a query image. 
We explore the latter direction, which is relatively underexplored compared to CIR, and leverage the joint embedding space of VLMs to effectively perform conditional image retrieval.
Concurrent work~\cite{liu2026conditional} also shares similar conditional representation learning via constructing the text-conditioned embedding space.
\vspace{-0.05em}
\section{\textcolor[HTML]{ff8b8b}{C}\textcolor[HTML]{f4ab44}{L}\textcolor[HTML]{74d674}{A}\textcolor[HTML]{9195fa}{Y}: Conditional Similarity Modulation}
\label{sec:method}
We propose a conditional visual similarity computation method leveraging VLMs~\cite{radford2021learning, zhai2023sigmoid} that consider hyperspherical manifold. We begin by describing the problem definition of conditional retrieval in Sec.~\ref{subsec:method_define}, followed by the formulation of our conditional similarity in Sec.~\ref{subsec:proposed_method}. We present the overall conditional similarity computation pipeline of our method in Fig.~\ref{fig:method}.

\subsection{Problem Definition}
\label{subsec:method_define}
Previous conditional retrieval methods~\cite{vaze2023genecis, hsieh2025focallens, meng2025vlm2vecv2,qwen3vlembedding} typically compute condition-based visual similarity by modifying image representations conditioned on text via a feature modulator.
In contrast, we reformulate the visual similarity space, allowing conditional retrieval with fixed embeddings.
We first categorize the conditional similarity computation methods into two formulations: symmetric and asymmetric methods based on how the text condition \textit{c} is incorporated into the image representations.
Given \textit{query} image $I_q$ and \textit{database} images $I_d$, the conditional similarity between $I_q$ and $I_d$ can be computed following these two formulations:
\begin{align}
\texttt{csim}_{\text{sym}}(I_q, I_d \mid c) &=
d\big(m(I_q, c),\, m(I_d, c)\big), \label{eq:sim} \\
\texttt{csim}_{\text{asym}}(I_q, I_d \mid c) &=
d\big(m(I_q, c),\, I_d\big), \label{eq:asim} 
\end{align}
where $d(\cdot, \cdot)$ denotes a similarity function such as cosine similarity.
The $m(I, c)$ represents a \textit{modulator} that integrates the input image $I$ and the given condition $c$.

\paragraph{Symmetric vs. asymmetric} Compared to the symmetric case~(Eq.~\ref{eq:sim}), the asymmetric computation~(Eq.~\ref{eq:asim}) does not forward the \textit{database} images through the modulator $m$ with the text condition.
A representative method GeneCIS~\cite{vaze2023genecis} follows the asymmetric formulation, where only \textit{query} features are conditioned on the text input.
However, despite the relative efficiency of this asymmetric formulation, the \textit{database} features will remain independent of the given condition.
Consequently, the retrieved results rely on the condition-agnostic representations, potentially leading to suboptimal retrieval performance when guiding the search.
To further support this, we experimentally compare the performance by applying both the symmetric and asymmetric formulations to 
conditional retrieval methods, in Sec~\ref{sec:experiments}.

\paragraph{\boldmath Design of modulator $m$} 
Existing methods~\cite{vaze2023genecis, hsieh2025focallens, meng2025vlm2vecv2,qwen3vlembedding} design the modulator $m$ through the neural network that incorporates text and image inputs.
In other words, the modulator is coupled with the visual feature extraction and the conditioning process through a neural network coupler to obtain conditional visual features.
This design requires a full forward pass through the network for each condition to obtain the corresponding conditional features.
In the symmetric form, it may lead to computational overhead and limit its practicality for adaptive conditional retrieval in a large-scale database.
In contrast, we decouple the conditioning process from the visual feature extraction within the modulator $m$, and propose a conditioning process that does not rely on the input visual feature.
We achieve this by proposing the conditional similarity space modulation scheme, which directly leverages the visual features from a pretrained VLM and projects them into a designed conditional similarity space.
This decoupling enables efficient adaptive retrieval in the symmetric form by eliminating the need to re-encode the database features for varying conditions.

\subsection{Visual Similarity Modulation}
\label{subsec:proposed_method}
As mentioned, since we utilize text modality for describing the condition, we adopt pretrained VLMs to leverage their structured joint embedding space.
VLMs typically consist of a vision encoder $f_I$ and text encoder $f_T$. 
For each \textit{query} $I_q$, \textit{database} image $I_d$ and condition $c$, we define $\mathbf{v}_q = f_I(I_q)$, $\mathbf{v}_d = f_I(I_d)$, and $\mathbf{t} = f_T(c)$, where $f_I$ and $f_T$ denote the vision and text encoder, respectively.

The main idea of conditional similarity space modulation is to project visual features onto the textual subspace, which captures the condition-aware relationships among visual features.
To achieve this, we derive a projection matrix $\mathbf{P}_c$ by performing singular value decomposition~(SVD) on the text condition feature matrix.
Specifically, following previous works~\cite{dorfman2025ip, nguyen2025csd}, we first generate condition-related textual prompts through Large-Language Model~(LLM) with the template: \texttt{a photo of} \{c\}. 
The generated text prompts are then encoded through a text encoder $f_T$ and their text embeddings $\mathbf{t}$ are concatenated to form the text feature matrix $\mathbf{T}_c = [\mathbf{t}_1^c, \dots, \mathbf{t}_n^c]^\top \in \mathbb{R}^{n \times d}$.
Previous works directly apply SVD on these text features to generate the projection matrix; however, this operation assumes Euclidean structure and therefore ignores the intrinsic geometry of the embedding manifold.
In contrast, since VLM embeddings lie on a unit hypersphere, this Euclidean assumption fails to capture their underlying geometry.

\paragraph{Manifold-aware textual subspace construction} To address this, we take into account the hyperspherical nature of the embedding manifold in VLMs to accurately model the relationship and derive a manifold-aware projection matrix $\mathbf{P}_c$.
We construct the textual subspace through a local tangent-space approximation rather than a Euclidean linear subspace, to alleviate the distortion from Euclidean projection similar to previous works~\cite{geodesic, berasi2025not}.
Under mild curvature assumptions, this local structure can be approximated by its tangent space at a reference point.
Previous works~\cite{liang2022mind, eslami2025mitigate} observe that the embedding geometry of each modality tends to be concentrated within a conical region, indicating that most embeddings lie closer to a common mean direction.
Hence, we leverage the validity of the local tangent space approximation under this geometric property.
Formally, any point $\mathbf{x} \in \mathcal{S}^{d-1} \setminus \{-\boldsymbol{\mu}\}$ can be mapped onto the tangent space
$\mathcal{T}_{\boldsymbol{\mu}} = \{\mathbf{x} \in \mathbb{R}^{d} : \mathbf{x}^\top \boldsymbol{\mu} = 0\}$ 
via \textit{logarithm map}~\cite{hauberg2018directional}:
\begin{equation}
\begin{gathered}
\log_{\boldsymbol{\mu}}(\mathbf{x}) := 
(\mathbf{x}-\boldsymbol{\mu}(\mathbf{x}^\top\boldsymbol{\mu}))
\frac{\theta}
{\sin(\theta)}, \\
\theta = \arccos({\mathbf{x}^\top\boldsymbol{\mu}}).
\end{gathered}
\end{equation}
We consider the normalized mean $\boldsymbol{\mu_c}$ of text features $\mathbf{T}_c$ as a reference point for defining the tangent space, and map text features from the unit hypersphere manifold to the tangent space of $\boldsymbol{\mu}_c$ with the \textit{logarithm} map.
Then, we apply SVD on these mapped text features:
\begin{equation}
\begin{gathered}
\log_{\boldsymbol{\mu}_c}({\mathbf{T}_c})
=\!\big[\!\log_{\boldsymbol{\mu}_c}(\mathbf{t}_1^c)\!
% \Log_{\boldsymbol{\mu}_c}(\mathbf{x}_2)\!
\cdots\!
\log_{\boldsymbol{\mu}_c}(\mathbf{t}_n^c)\!\big]^\top, \\
\log_{\boldsymbol{\mu}_c}(\mathbf{T}_c) = \mathbf{U}\boldsymbol{\Sigma} \mathbf{V}^\top.
\end{gathered}
\end{equation}
We utilize the top-$k$ right singular vectors to construct textual subspace, 
% defined as $\text{span}\{\mathbf{v}_1,\dots,\mathbf{v}_k\}$, 
and obtain the projection matrix $\mathbf{P}_c = {\mathbf{V}_k}{\mathbf{V}_k}^\top$.
By precomputing these textual subspaces for each condition (\ie, projection matrices), we only need to modulate the conditional similarity space using precomputed projection matrices at inference, accordingly.
Specifically, we project the query and database visual features from pre-trained VLMs into the space that satisfies the desired conditional relationships, without re-encoding.

\paragraph{Inference}
Now we have a projection matrix $\mathbf{P}_c$ that projects features onto the textual subspace within the tangent space of $\boldsymbol{\mu}_c$.
Our ultimate goal is to model the intra-relationship among visual features within this condition-aware textual subspace.
As mentioned above, the \textit{logarithm map} onto the tangent space is valid under mild curvature assumptions.
However, due to the conic effect~\cite{liang2022mind, eslami2025mitigate}, naive projection of visual features cannot maintain this, which means that visual features are distant from reference point.

To mitigate this, we apply an orthonormal rotation $H(\cdot)$ to align the mean of the database visual features $\boldsymbol{\mu}_{v_d}$ with the mean of text features $\boldsymbol{\mu}_c$, without altering the intra-relationship among the visual features. By aligning the visual features with the textual mean, this operation enables tangent space mapping for both modalities to be performed at the consistent reference point.
We then apply the logarithm map to map the rotated visual features $H(\mathbf{v})$ onto the tangent space at $\boldsymbol{\mu}_c$, and subsequently project them onto the textual subspace with projection matrix $\mathbf{P}_c$.
Finally, similar to standard similarity computation, we compute the similarity between the query feature $\mathbf{v}_q=f_I(I_q)$ and the database features $\mathbf{v}_d=f_I(I_d)$ via cosine similarity.
The proposed conditional similarity between the query image $I_q$ and the database image $I_d$ under the condition $c$ is as follows:
\begin{equation}
\begin{gathered}
m_{\text{CLAY}}(I, c) := \mathbf{P}_c\log_{\boldsymbol{\mu}_c}\big(H(f_I(I))\big), \\
\mathtt{csim}_{\text{CLAY}} (I_q, I_d \mid c) := d\big(m_{\text{CLAY}}(I_q, c), m_{\text{CLAY}}(I_d, c)\big),
\end{gathered}
\end{equation}
where $d(\cdot, \cdot)$ corresponds to the cosine similarity. The detailed pipeline is provided in the supplementary material.

\section{Our CLAY-EVAL Dataset}
\label{sec:synthetic_dataset}
Synthetic benchmark datasets have emerged as a powerful tool for providing scalable, controlled evaluations.
Leveraging the high fidelity of recent generative models, we provide a novel synthetic evaluation dataset, CLAY-EVAL, for conditional image retrieval.
To evaluate our context-aware conditional image retrieval setting, each image in the dataset should be annotated with multiple labels, allowing flexible clustering under different conditions.
While several datasets are classified with particular conditions, their scalability is limited because real-world images were manually annotated~\cite{kwon2024ictc}, they only have a single condition~\cite{yao2011human, Nilsback08, parkhi12a, wang2010locality, yao2010grouplet}, or they include only simple simulated 3D geometric objects (\eg, cone, sphere, and cube) with a simple background~\cite{johnson2017clevr}.
To address this, we further build a synthetic dataset using a powerful open-source diffusion model~\cite{Flux2024}, whose reliability allows for the controlled generation of high-quality and diverse image samples.

\begin{figure}[t]
    \centering  
   \includegraphics[width=0.90\linewidth]{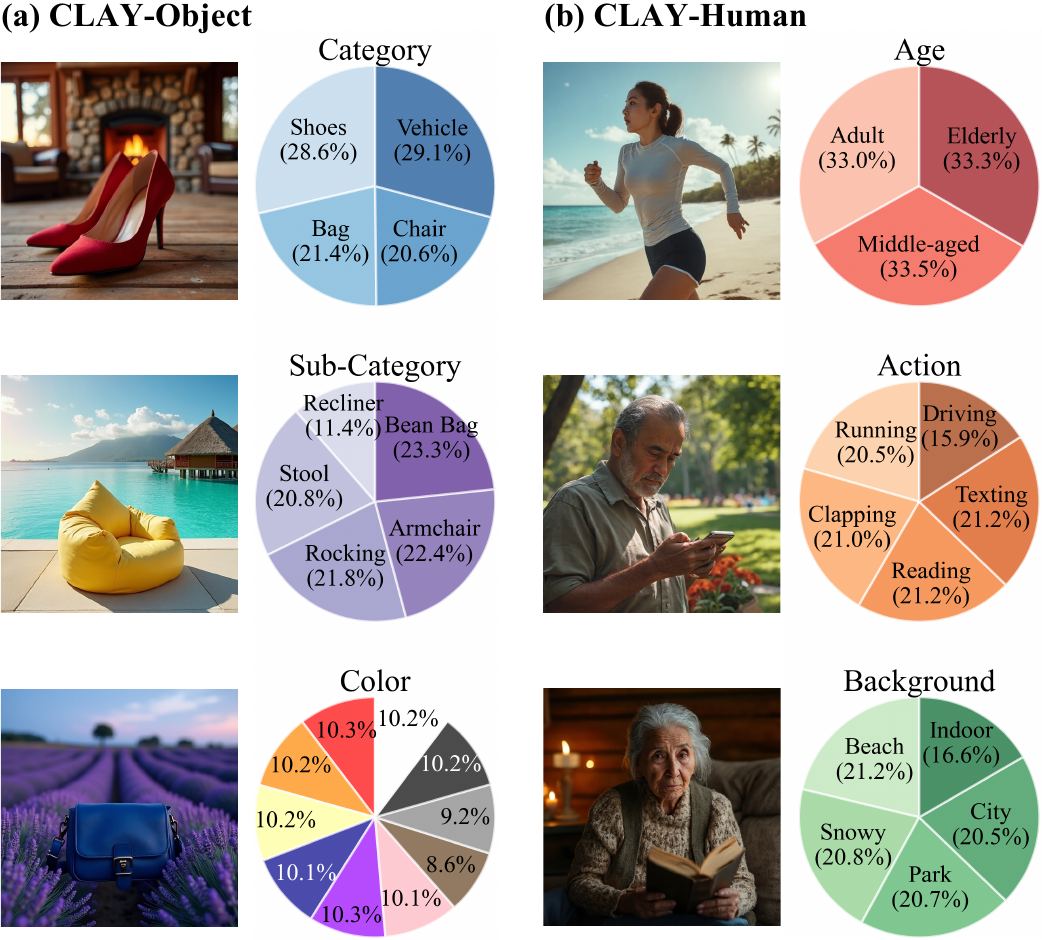}
    \caption{\textbf{Our CLAY-EVAL dataset statistics.} We construct a synthetic dataset with diverse condition annotations, consisting of (a) \textit{object} entity and (b) \textit{human} entity. For both, the left column shows sample images demonstrating visual naturalness, and the right column visualizes the distributions of key attributes, showing diversity. Percentages are truncated to one decimal place and annotation text labels are abbreviated.
    }
\label{fig:dataset_statistics}
\vspace{-1.5em}
\end{figure}

To construct an evaluation dataset aligned with our objectives,
we establish three key design principles: disentanglement, compositionality, and naturalness, inspired by ~\citet{li2024omniprism}.
Based on these, we structure our dataset into two main entities, \textit{object} and \textit{human}. 
Each dataset consists of core attributes, which serve as the primary query conditions, and diversity attributes, which control for visual variance and bias.
We define three core attributes for each entity, namely category, sub-category, and color for the \textit{object} entity, and age, action and background for the \textit{human} entity.

We generate images with structured prompts derived from all attribute combinations.
For stringent quality control, we first exclude contradictory combinations via schema-level filtering, and subsequently manually remove generated images exhibiting low text-visual alignment. As shown in Fig.~\ref{fig:dataset_statistics}, our synthetic dataset covers a wide range of instances and paired images, enabling comprehensive context-aware conditional retrieval evaluation. 
Our final synthetic dataset consists of 7,325 \textit{object} images and 6,745 \textit{human} images, and
detailed lists of all attribute instances and curation process are provided in the supplementary material.

\section{Experiments}
\label{sec:experiments}
\begin{table*}[t] 
\caption{\textbf{Quantitative comparison of mean Average Precision~(mAP) on single conditional datasets.}
We evaluate single conditional retrieval performance across various datasets, where each condition corresponds to a specific attribute (\eg, \texttt{action}, \texttt{cat species}, \texttt{flower type}).
(a) Real-world datasets and (b) synthetic datasets are presented separately.
We compare the baseline Vision-Language Models~(VLMs) and competing methods that support conditional retrieval.
The \colorbox{blizzardblue}{best} and \colorbox{bubbles}{second-best} results are highlighted.}
\centering 
\begin{subtable}[t]{\textwidth} 
\footnotesize
\renewcommand{\figurename}{Table} 
\resizebox{0.99\textwidth}{!}{
\renewcommand{\arraystretch}{1.0}
\begin{tabular}{m{3cm} cccccccccc} 
\toprule 
\multirow{2}[2]{*}{\textbf{Method}} & \multicolumn{3}{c}{\textbf{Stanford40}} & \multicolumn{7}{c}{\textbf{Fine-grained Image Classification}} \\ \cmidrule(lr) {2-4} \cmidrule(lr) {5-11}
& \textbf{Action} & \textbf{Location} & \textbf{Mood} & \textbf{Cat} & \textbf{Dog} & \textbf{Instrument} & \textbf{Flower} & \textbf{Car} & \textbf{Aircraft} & \textbf{Food} \\ \midrule
CLIP-B & 43.0&  47.0&  53.0& 37.5& 37.9& 27.1& 70.1& 30.5& 20.9& 47.4\\
SigLIP-B  & 54.8& 52.7& 56.4& 55.9& 59.3& 40.6& \cellcolor{bubbles}{86.6}& \cellcolor{bubbles}{64.6}& \cellcolor{bubbles}{46.7}& \cellcolor{bubbles} {61.3}\\
\midrule
GeneCIS & 50.0& 50.9& 51.8&24.7& 24.7&24.7&58.5&  16.9& 17.2& 44.8\\
SEARLE & 40.5 & 43.2 & 50.0 & 32.6 & 43.5 & 20.4 & 34.8 & 17.2 & 11.0  & 31.1 \\
MagicLens & 52.6 & 47.5 & 55.4 & 29.1 & 41.7 & 23.8 & 47.9 & 16.5 & 9.3 & 41.9 \\
InstructBLIP  & 63.1& 54.4& \cellcolor{blizzardblue} {60.3}& 50.6& 56.8& 37.1& 76.5& 27.6& 15.7& 50.6\\
Ours (CLIP-B) & \cellcolor{bubbles}{66.0}& \cellcolor{bubbles}{55.4}& 57.9& \cellcolor{bubbles}{72.7}& \cellcolor{bubbles}{79.3}& \cellcolor{bubbles}{58.0}& 80.4& 51.1& 28.4& 58.8\\
Ours (SigLIP-B) & \cellcolor{blizzardblue} {66.2}& \cellcolor{blizzardblue}{59.5}& \cellcolor{bubbles}{58.7}& \cellcolor{blizzardblue}{ 82.1}& \cellcolor{blizzardblue}{84.7}& \cellcolor{blizzardblue}{63.4}& \cellcolor{blizzardblue}{92.7}& \cellcolor{blizzardblue}{78.0}& \cellcolor{blizzardblue}{57.9}& \cellcolor{blizzardblue}{66.1}\\
\bottomrule
\end{tabular}
}%
\captionsetup{font=small} \caption{{real-world datasets}.} 
\label{tab:single_real}
\end{subtable}

\begin{subtable}[t]{\textwidth} 
\footnotesize
\renewcommand{\figurename}{Table} 
\resizebox{0.99\textwidth}{!}{
\renewcommand{\arraystretch}{1.0}
\begin{tabular}{m{2.5cm} cccccccccc} 
\toprule 
\multirow{2}[2]{*}{\textbf{Method}}
& \multicolumn{4}{c}{\textbf{Clevr4}} 
& \multicolumn{3}{c}{\textbf{CLAY-Object}} & \multicolumn{3}{c}{\textbf{CLAY-Human}} \\
\cmidrule(lr) {2-5} \cmidrule(lr) {6-8} \cmidrule(lr) {9-11}
& \textbf{Shape} & \textbf{Color} & \textbf{Texture} & \textbf{Count} 
& \textbf{Color} & \textbf{Category} & \textbf{Subcategory}  
& \textbf{Age} & \textbf{Action} & \textbf{Background} \\
\midrule
CLIP-B & 61.7 & 19.9 & 18.1 & 13.9 & 12.9 & 68.9& 42.4&  50.4& 54.4& 41.8\\
SigLIP-B  & 77.7 & 20.0 & 18.6 & 13.0 & 14.7 & 69.6& 62.0& 47.7& 56.7& 53.6\\
\midrule
GeneCIS & 47.9 & 16.6 & 16.0 & 11.5& 12.5 & 80.1& 38.4& 45.1& 62.4& 46.2\\
SEARLE & 43.0   & 20.7 & 13.8 & 13.0   & 12.1 & 69.1 & 38.8 & 46.1 & 53.7 & 40.1 \\
MagicLens & 45.1 & 21.2 & 12.9 & 13.7 & 17.3 & 59.9 & 39.3 & 43.4 & 62.6 & 48.3 \\
InstructBLIP  & \cellcolor{bubbles}83.4 & 27.4 & \cellcolor{bubbles}22.8 &17.9 & 14.4 & 75.6& 61.4& 45.0& 72.2& 73.7\\
Ours (CLIP-B) & 72.1 & \cellcolor{bubbles}67.9 & 22.2 & \cellcolor{bubbles}21.8 & \cellcolor{bubbles}47.8 & \cellcolor{bubbles}93.4& \cellcolor{bubbles}65.0& \cellcolor{blizzardblue}71.3& \cellcolor{blizzardblue}81.3& \cellcolor{bubbles}82.4\\
Ours (SigLIP-B) & \cellcolor{blizzardblue}88.6 & \cellcolor{blizzardblue}73.2 & \cellcolor{blizzardblue}26.1 & \cellcolor{blizzardblue}22.7 & \cellcolor{blizzardblue}63.3& \cellcolor{blizzardblue}94.3& \cellcolor{blizzardblue}81.9& \cellcolor{bubbles}61.0& \cellcolor{bubbles}80.4& \cellcolor{blizzardblue}89.4\\
\bottomrule
\end{tabular}
}%
\captionsetup{font=small} \caption{{synthetic datasets}.} 
\label{tab:single_syn}
\end{subtable}
\vspace{-2em}
\label{tab:single}
\end{table*}

\subsection{Experimental Setup}
\noindent\textbf{Evaluation datasets and metrics.}
We evaluate the conditional retrieval performance on a wide range of datasets, including both real-world datasets and our synthetic dataset.
For real-world datasets, we utilize fine-grained image classification datasets~\cite{Nilsback08, Krause_2013_stanfordcars, maji2013aircraft, bossard2014food,parkhi12a, wang2010locality, yao2010grouplet},
and further conduct experiments on Stanford40~\cite{yao2011human} with the human annotated labels from~\cite{kwon2024ictc}.
In these fine-grained classification datasets, we consider each category as condition, for example, in Stanford40, the condition we suppose is \texttt{action}.
For the synthetic evaluation, we employ CLEVR4~\cite{johnson2017clevr} with the conditions \texttt{shape}, \texttt{color}, \texttt{texture}, and \texttt{count}, as well as our generated synthetic dataset.
We split each dataset into \textit{query} and \textit{database} with a 1:9 ratio, and in the evaluation with \texttt{subcategory} condition, we report the averaged performance from each categorized database.
For the evaluation metric, we adopt the standard metric mean Average Precision~(mAP) following conventional image retrieval tasks~\cite{gordo2017end, radenovic2018revisiting, revaud2019learning}, unless otherwise noted.

\begin{table}[t] 
\centering 
\caption{\textbf{Quantitative comparison of mAP on multi-conditional datasets CLAY.} We provide the condition combinations such as \texttt{color and category}. We report mAP as an evaluation metric. The best results are highlighted.}
\resizebox{1\linewidth}{!}{
\renewcommand{\arraystretch}{1.2}
\begin{tabular}{m{2.5cm} cc cccc} 
\toprule 
\multirow{2}{*}{\textbf{Method}} 
& \multicolumn{2}{c}{\textbf{CLAY-Object}} 
& \multicolumn{4}{c}{\textbf{CLAY-Human}} \\ 
\cmidrule(lr){2-3} \cmidrule(lr){4-7}

& \makecell{\textbf{Color} \\ \textbf{Category}}
& \makecell{\textbf{Color} \\ \textbf{Subcategory}}
& \makecell{\textbf{Age} \\ \textbf{Action}}
& \makecell{\textbf{Age} \\ \textbf{Background}}
& \makecell{\textbf{Action} \\ \textbf{Background}}
& \textbf{All} \\

\midrule
CLIP-B & 11.6 & 10.9 & 35.9 & 27.1 & 37.0 & 31.9 \\
SigLIP-B & 13.9 & 19.7 & 34.5 & 32.1 & 50.5 & 39.5 \\
\midrule
InstructBLIP  & 12.5 & 14.7 & 38.6 & 31.2 & 74.7 & 44.1 \\
Ours (CLIP-B) & 35.9 & 38.2 & \textbf{57.0} & \textbf{56.6} & 78.6 & \textbf{58.0} \\
Ours (SigLIP-B) & \textbf{44.7} & \textbf{55.0} & 49.0 & 55.6 & \textbf{81.5} & 52.0 \\
\bottomrule
\end{tabular}
}
\vspace{-1em}
\label{tab:multi}
\end{table}

\paragraph{Competing methods}
To the best of our knowledge, GeneCIS~\cite{vaze2023genecis} is the only work that is closely related to our problem.
% As mentioned above, they follow the asymmetric formulation, only conditioning the \textit{query} features.
% We further apply GeneCIS in symmetric formulation by also feed-forwarding the \textit{database} features with the conditions through the modulator, noted as GeneCIS$^{\mathbf{\dagger}}$.
For a more comprehensive comparison, we include Composed Image Retrieval methods~\cite{baldrati2023zero,zhang2024magiclens} and InstructBLIP~\cite{dai2023instructblip}.
To extract the conditioned visual features in InstructBLIP, we average Q-Former output tokens, following FocalLens~\cite{hsieh2025focallens}.
We further evaluate multi-modal LLM embedding models such as Qwen3-VL-Embedding~\cite{qwen3vlembedding} and VLM2Vec~\cite{meng2025vlm2vecv2} to compare different similarity computation formulations.
Detailed condition instructions and implementation details are provided in the supplementary materials.

\paragraph{Implementation details}
In our method, we utilize two VLMs, CLIP~\cite{radford2021learning} and SigLIP~\cite{zhai2023sigmoid}.
To generate condition-related textual prompts, we utilize ChatGPT-5. We truncate the top-$k$ singular vectors and select $k=50$ for all experiments.
Additional details are in the supplementary materials.

\begin{figure*}[t]
\centering
\captionsetup{type=figure}
\includegraphics[width=0.95\linewidth]{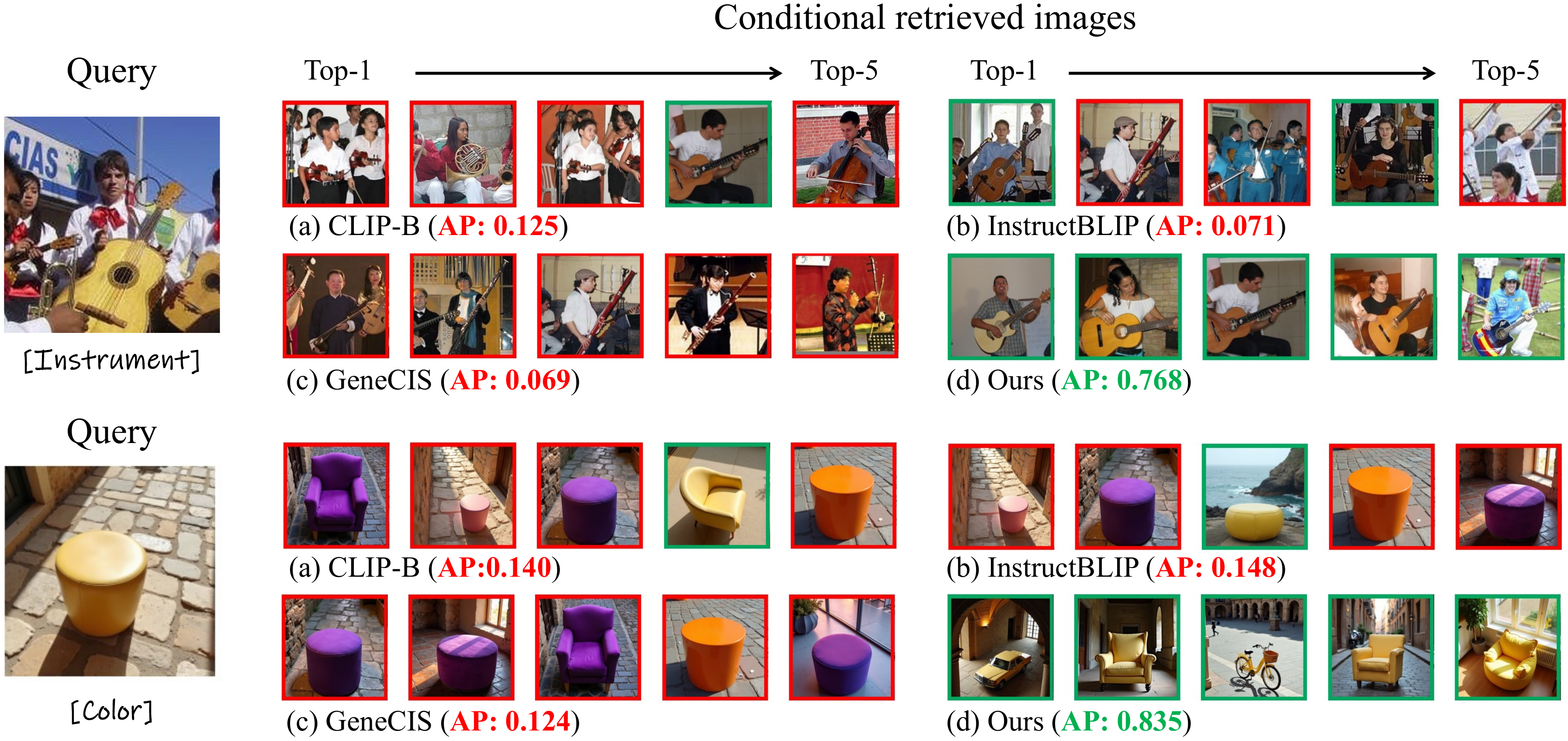}
    \vspace{-1.3mm}
   \caption{\textbf{Qualitative comparison of our method with competing methods.} For each query image and condition text pair, we compare the top-5 retrieved results from (a) CLIP-B, (b) InstructBLIP, (c) GeneCIS, and (d) our method. We also report Average Precision~(AP) in each retrieval result. \textcolor[HTML]{00a65d}{Green} boxes indicate correctly retrieved images, while incorrect retrievals are shown in \red{red} boxes.
   }
\label{fig:qual_1}
\end{figure*}

\begin{figure}[t]
\centering
\captionsetup{type=figure}
\includegraphics[width=0.97\linewidth]{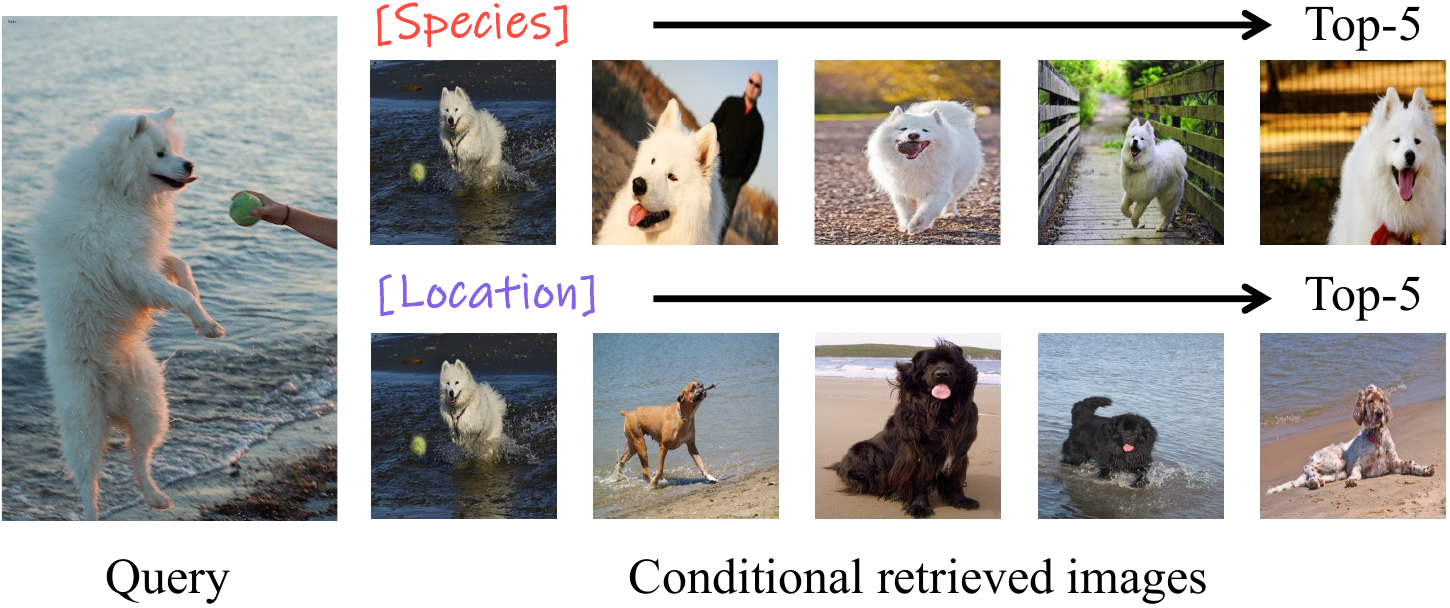}
    \vspace{-1.3mm}
   \caption{\textbf{Qualitative result on Oxfordpets dataset.} We visualize the top-5 retrieved results with condition \texttt{dog species} and \texttt{location}. Since no ground-truth location labels are available, we present qualitative examples only.} 
\label{fig:qual_2}
\vspace{-1.5em}
\end{figure}

\subsection{Comparison of Retrieval Performance}
\begin{figure*}[!htp]
\centering
\captionsetup{type=figure}
\includegraphics[width=0.99\linewidth]{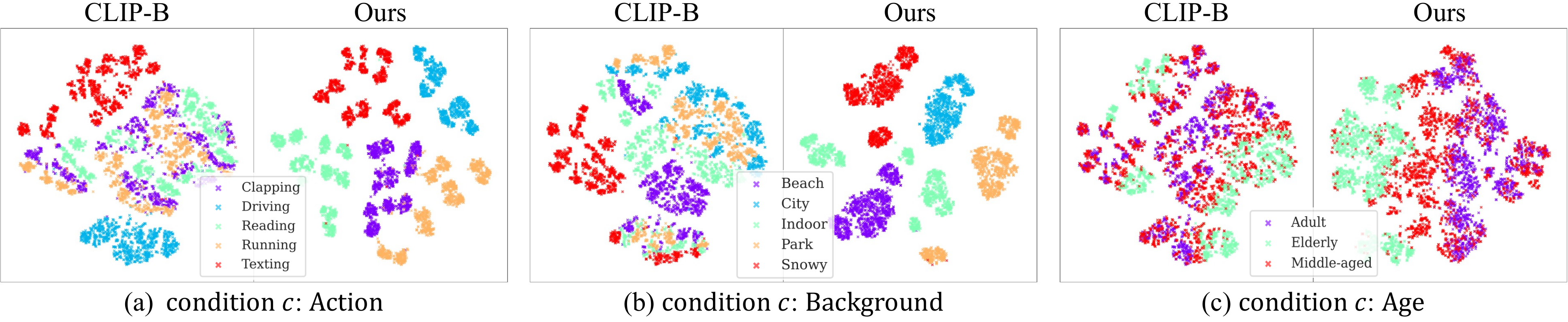}
    \vspace{-1.3mm}
   \caption{\textbf{Representation space visualization with t-SNE.} We report t-SNE of CLIP-B and ours~(CLIP-B) on CLAY-Human under condition (a) \texttt{action}, (b) \texttt{background}, and (c) \texttt{age}. The features with the same label are shown in the same color for easy interpretation. Compared to the fixed representation space in CLIP-B, our method forms more discriminative spaces compliant with given conditions.
   } 
\label{fig:tsne}
\end{figure*}

Table~\ref{tab:single} presents the single conditional retrieval accuracy of our method and competing methods across various real and synthetic evaluation datasets.
For clarity, we also report the baseline performance of CLIP-B and SigLIP-B, which cannot input text conditions. As shown in this table, CLAY consistently outperforms competing methods, exhibiting strong adaptability across diverse datasets and conditions, highlighting that similarity modulation can yield highly effective results.
In contrast, our method inherits the strong zero-shot capability of pre-trained VLMs, achieving state-of-the-art performance across diverse datasets and varying condition types without incurring high computational overhead in the symmetric setting.
In addition, CLAY can easily be extended to multi-conditional retrieval by constructing text feature matrix from multiple condition-related textual prompts.
Table~\ref{tab:multi} indicates CLAY performs reliably under the multi-conditional retrieval setting, achieving high accuracy on CLAY-Object and CLAY-Human datasets.
% We further present the quantitative results on GeneCIS benchmark ``Focus Attribute'' subset, compared to Composed Image Retrieval methods~\cite{zhang2024magiclens, karthik2024vision, baldrati2023zero}. Table~\ref{tab:genecis} shows CLAY still achieves the best performance when utilizing the same or even smaller backbone (\ie, ViT-B).
In Fig.~\ref{fig:qual_1}, we visualize conditional retrieved results, and it demonstrates that our method achieves better condition-aligned retrieval than the comparison methods.
Furthermore, as shown in Fig.~\ref{fig:qual_2}, CLAY effectively reflects diverse conditions within the same query.
Figure~\ref{fig:tsne} also shows how the representation space varies under different conditions.
This wide range of coverage is crucial in real-world scenarios, where users may have various intentions that need to be satisfied.
\begin{table}[t]
\caption{\textbf{Comparison of retrieval performance and time between symmetric, asymmetric similarity formulations.
} We evaluate retrieval accuracy and time efficiency of the two formulations. We report the average latency for a single query under the first condition and the second condition. $^{\mathbf{\dagger}}$ denotes the symmetric formulation.
} \label{tab:mAP-time_comp}
\vspace{-0.5em}
\renewcommand{\arraystretch}{1}
\resizebox{1\linewidth}{!}{
\begin{tabular}{l ccccc cc} \toprule
\multirow{2}{*}{\textbf{Method}} & \multicolumn{5}{c}{\textbf{mAP ($\uparrow$)}} & \multicolumn{2}{c}{\textbf{Time (s) ($\downarrow$)}} \\ 
\cmidrule(lr){2-6} \cmidrule(lr){7-8} 
 & \textbf{Stanford} & \textbf{Fine-g.} & \textbf{Clevr} & \textbf{Obj.} & \textbf{Human} & \textbf{1st} & \textbf{2nd} \\ \midrule
GeneCIS          & 50.9 & 30.2 & 23.1 & 43.7 & 51.2 & 1.705 & 0.041\\
Qwen3-VL     & 63.4 & 68.6 & 43.7 & 66.8 & 72.8 & 12.73 & 0.191\\ \midrule
GeneCIS$^{\mathbf{\dagger}}$          & 57.7 & 40.7 & 30.2 & 45.3 & 54.8 & 1.705 & 1.657 \\
Qwen3-VL$^{\mathbf{\dagger}}$     & 76.0 & 85.3 & 60.2 & 75.5 & 85.8 & 12.74 & 12.74 \\
VLM2Vec$^{\mathbf{\dagger}}$          & 73.1 & 81.1 & 71.7 & 73.4 & 88.7 & 13.61 & 13.63\\ \midrule
Ours (CLIP-L)    & 62.5 & 72.5 & 47.3 & 71   & 77.7 & 2.206 & 0.092\\
Ours (SigLIP-L)  & 64.7 & 84.3 & 53.0 & 80.7 & 79.7 &2.250 & 0.126\\ \bottomrule
\end{tabular} 
}
\vspace{-1em} 
\end{table}

\subsection{Comparison of Similarity Formulations}
In Sec.~\ref{sec:method}, we categorize the conditional similarity computation methods into asymmetric and symmetric formulations. The asymmetric form may lead to suboptimal performance because database features remain condition-agnostic, whereas the symmetric form incurs high computational overhead due to the database modulation process.
To validate this, we compare the two formulations by applying both to GeneCIS~\cite{vaze2023genecis} and Qwen3-VL-Embedding~\cite{qwen3vlembedding}, and additionally report results from VLM2Vec~\cite{meng2025vlm2vecv2}.
We report the averaged retrieval accuracy across datasets and the averaged inference time per query, measuring separately for the first condition and for the subsequent second condition.
For a controlled evaluation of inference time, we sample 100 images from CLAY-Object for the database and report the average retrieval time per query across 10 runs.

Specifically, since GeneCIS follows the asymmetric formulation only conditioning the \textit{query} features, we apply the symmetric formulation by feed-forwarding the \textit{database} features with the conditions through the modulator.
As shown in Tab.~\ref{tab:mAP-time_comp}, compared to the asymmetric case, the symmetric formulation consistently improves retrieval accuracy in both GeneCIS and Qwen3-VL-Embedding. This result indicates that the conditioning process for both query and database is critical for capturing condition-specific relationships.
However, this performance gain comes with a high computational overhead, since applying a new condition requires recomputing the embeddings of the whole database, which makes it impractical for large-scale database scenarios.
In contrast, our method effectively balances the strength of symmetric and asymmetric paradigms. CLAY achieves strong retrieval performance without additional feature encoding cost, placing it in a practical balance.

\subsection{Analysis of Representation Space}
To confirm the effectiveness of similarity space modulation,
we visualize t-SNE results comparing with CLIP-B and Ours (CLIP-B) on our CLAY-Human dataset.
As shown in Fig~\ref{fig:tsne}, the visual representation space is distinctly separated with each \texttt{action} and \texttt{background} condition, compared to the baseline.
This result shows how our method adaptively modulates the original visual features onto the condition-aware subspace, achieving high conditional retrieval performance.
Interestingly, applying our method shows a rankable property along the condition axis (\ie, \texttt{age}) in Fig~\ref{fig:tsne} (c).
Previous work~\cite{sonthalia2025rankability} studied this rankability by identifying the rankable axis,  CLAY can achieve this property naturally through conditional similarity space modulation.
We also find that this rankable property holds in Clevr4-count, and provide in supplementary materials.

\begin{table}[t]
\caption{\textbf{Effectiveness of rotation and manifold modeling.} We ablate the rotation operation and manifold modeling across datasets.
We report the averaged mAP. Due to space limitations, we abbreviate manifold modeling as manifold, and CLAY-Human and CLAY-Object as Human and Object, respectively.}
\centering
\renewcommand{\arraystretch}{1.3} 
    \resizebox{1\linewidth}{!}{
    \begin{tabular}{lcccccc} \toprule
    \textbf{Rotation} & \textbf{Manifold} & \textbf{Stanford40} & \textbf{Fine-grained} & \textbf{Clevr4} & \textbf{Object} & \textbf{Human} \\ \midrule
    \textcolor{red}{\xmark} & \textcolor{red}{\xmark} & 57.9 & 60.5 & 43.8 & 65.4 & 76.8 \\
    \textcolor{red}{\xmark} & \textcolor{green(ncs)}{\cmark} & 57.9 & 60.5 & 43.9 & 63.7 & 76.8 \\
    \textcolor{green(ncs)}{\cmark} & \textcolor{green(ncs)}{\cmark} & \textbf{59.8} & \textbf{61.3} & \textbf{46.0} & \textbf{66.9}   & \textbf{78.3} \\ \bottomrule
    \end{tabular}}
\label{tab:ablation}
\vspace{-1em}
\end{table}
\subsection{Ablation Study}
Table~\ref{tab:ablation} shows the effects of rotation and manifold modeling on retrieval performance. 
We find that applying manifold-aware modeling does not provide meaningful improvements or even degraded performance on specific datasets.
This implies that simply considering the hyperspherical geometry of embedding space is not sufficient since the visual features are not well aligned with the reference point.
In contrast, when the rotation operation is combined with the manifold modeling, the performance improves across datasets, indicating that the rotation step is important for effectively preserving relationships during similarity space modulation.

\section{Conclusion}
\label{sec:conclusion}
In this work, we introduce CLAY, a novel conditional visual similarity computation method that adaptively modulates the fixed similarity of existing pre-trained VLMs to a text-conditional similarity without additional training.
CLAY's adaptive conditioning strikes a sweet spot between accuracy and efficiency that prior methods struggle to achieve simultaneously, thereby enabling practical and effective conditional retrieval.
By leveraging the underlying geometry, our manifold-aware textual subspace enables theoretically grounded modeling of the conditional relationship between images.
To support a comprehensive evaluation of the multi-faceted aspects of conditional retrieval, we construct a synthetic evaluation dataset containing diverse object and human images annotated with conceptual conditions.
We believe this work opens up promising directions for building practical retrieval systems compliant with human intentions as well as potential applications beyond retrieval systems, \eg, text and visual matching in multimodal generative models \cite{dat2025vsc} by introducing a focused similarity like ours.
\clearpage
\resumetocwriting
\noindent\textbf{Acknowledgments.}
This work was supported by the InnoCORE program of the Ministry of Science and ICT(N10250156, KAIST InnoCore LLM; 25-InnoCORE-01, Trust-Enhanced Mutualistic Bio-Embedded AI),
Samsung Electronics Co., Ltd (Project Code: IO260114-15161-01),
Institute of Information \& communications Technology Planning \& Evaluation (IITP) grant funded by the Korea government(MSIT) (No. RS-2024-00457882, National AI Research Lab Project; No. 2022-0-00124, No. RS-2022-II220124, Development of Artificial Intelligence Technology for Self-Improving Competency-Aware Learning Capabilities).

{
    \small
    \bibliographystyle{ieeenat_fullname}
    \bibliography{ref}
}

\clearpage
\maketitlesupplementary
\tableofcontents
\noindent\rule{\linewidth}{0.5pt}
\setcounter{page}{1}
\setcounter{section}{0}
\renewcommand{\thefigure}{S\arabic{figure}}
\renewcommand{\thetable}{S\arabic{table}}
\setcounter{figure}{0} 
\setcounter{table}{0} 
\renewcommand{\thesection}{\Alph{section}}

In this supplementary material, we provide more details and results which are not included in the main paper due to space constraints. 
Sec.~\ref{sec:supp_clay_details} details of our method with pseudo code.
In Sec.~\ref{sec:supp_clay_eval}, we provide the construction process of our synthetic dataset CLAY-EVAL.
Sec.~\ref{sec:supp_exp_details} offers experimental details including setup and implementations.
We provide additional results of ablation, qualitative, t-SNE in Sec.~\ref{sec:supp_additional_results}.
Finally, we discuss limitations of our method in Sec.~\ref{sec:supp_limitations}.
We further kindly encourage readers to see the supplementary HTML file containing additional qualitative results.

\section{Method Details}
\label{sec:supp_clay_details}
In this section, we provide a detailed explanation of our method.
Our method consists of two main processes: (1) construction of projection matrix and (2) computation of visual similarity between query and database images.

\begin{table*}[t]
\centering
\caption{\textbf{LLM Prompts for generating condition-related text descriptions.}}
\begin{adjustbox}{}
\begin{tabular}{p{0.9\textwidth}}
\toprule
\textbf{LLM system prompts.} \\
\midrule
You will be given a user prompt. Your job is to determine the user intent (where to focus on in the image) and generate 100 prompts related to the intent, following the format provided. \\
\\
For example, given the following user intent: ``I want to focus on the human action in this image.'' \\
An exemplar answer is: \\
\\
Intent: action  
Prompts: a photo of running, a photo of jumping, a photo of swimming, ... \\
\\
Output must always be in the exact format: \\
Intent: $<$intent-word$>$ \\
Prompts: $<$string$>$, $<$string$>$, ..., $<$string$>$ \\
- The Prompts list must contain exactly 100 unique entries following the format ``a photo of \{intent\}''. \\
- Each string should directly replace the placeholder (e.g., {human action}) with a real example without inserting any other phrases. \\
- Do not include numbering or bullet points or extra quotes.
\\ \midrule
\textbf{Generation example.} \\
\midrule
LLM input : I want to focus on the human action in this image \\
LLM output: \\
Intent: action \\
Prompts: a photo of running, a photo of jumping, a photo of swimming, a photo of dancing, ... \\
\bottomrule
\end{tabular}
\end{adjustbox}
\label{tab:supp_llm_prompts}
\end{table*}

\paragraph{Construction of projection matrix}
As presented in the main paper, we first generate condition-related text prompts with Large Language Model~(LLM).
We automatically construct a system prompt template with LLM in advance, to effectively filter non-related generations and ensure the response template, as shown in Tab.~\ref{tab:supp_llm_prompts}.
Then, we input the LLM to generate texts with given condition $c$, followed by forwarding step to the text encoder of VLM:
\begin{align}
    c_i &\in \mathcal{C}, \quad i = 1,\ldots,n, \\
    \mathbf{t}_i^c &= f_T(c_i), \\
    \mathbf{T}_c &= [\mathbf{t}_1^c, \mathbf{t}_2^c, \dots, \mathbf{t}_n^c]^\top ,
\end{align}
where $c_i$ is the $i$-th LLM generated text prompt.
As in Sec.~4, we map the features onto the tangent space before applying Singular Value Decomposition~(SVD).
We choose the normalized mean of text features as a reference point, and define \textit{logarithm map} to map the points lying on the hyperspherical manifold onto the tangent space of $\boldsymbol{\mu}_c$ following~\cite{hauberg2018directional, geodesic, berasi2025not}:
\begin{align}
    \boldsymbol{\mu}_c = \frac{\frac{1}{n}\sum_{i=1}^{n} \mathbf{t}_i}{\left\lVert \frac{1}{n}\sum_{i=1}^{n} \mathbf{t}_i \right\rVert},
\end{align}
\vspace{-0.5em}
\begin{equation}
\begin{gathered}
    \log_{\boldsymbol{\mu}_c}(\mathbf{x}) := 
    (\mathbf{x}-\boldsymbol{\mu}_c(\mathbf{x}^\top\boldsymbol{\mu}_c))
    \frac{\theta}
    {\sin(\theta)}, \\
    \theta = \arccos({\mathbf{x}^\top\boldsymbol{\mu}_c}).
\end{gathered}
\label{eq:supp_log_map}
\end{equation}
After we map the features onto the tangent space with Eq.~\ref{eq:supp_log_map}, we then generate projection matrix $\mathbf{P}_c$ by applying SVD:
\begin{equation}
\begin{gathered}
\log_{\boldsymbol{\mu}_c}({\mathbf{T}_c})
=\!\big[\!\log_{\boldsymbol{\mu}_c}(\mathbf{t}_1^c),\!
\cdots\!
,\log_{\boldsymbol{\mu}_c}(\mathbf{t}_n^c)\!\big]^\top, \\
\log_{\boldsymbol{\mu}_c}(\mathbf{T}_c) = \mathbf{U}\boldsymbol{\Sigma} \mathbf{V}^\top, \\
\end{gathered}
\end{equation}
\vspace{-1.5em}
\begin{align}
\mathbf{P}_c = \mathbf{V}_k\mathbf{V}_k^\top.
\end{align}

\paragraph{Computation of visual similarity}
To alleviate the approximation error using the \textit{logarithm map} resulting from the gap between text-visual features, also known as \textit{conic effect}~\cite{liang2022mind}, we first mitigate the gap by aligning the mean of visual features with $\boldsymbol{\mu}_c$.
To achieve this, we apply two householder transformations so that the distances and angles between two visual features remain unchanged, while the visual mean becomes aligned with the textual mean:
\begin{align}
\boldsymbol{\tilde{\mu}} = \frac{\boldsymbol{\mu}_v + \boldsymbol{\mu}_c}{\|\boldsymbol{\mu}_v + \boldsymbol{\mu}_c\|},
\end{align}
\begin{equation}
\begin{gathered}
\mathbf{H}_1=\mathbf{I} - 2\,\frac{(\boldsymbol{\mu}_v - \boldsymbol{\tilde{\mu}})(\boldsymbol{\mu}_v-\boldsymbol{\tilde{\mu}})^{\top}}{\left\|\boldsymbol{\mu}_v - \boldsymbol{\tilde{\mu}}\right\|^{2}}, \\
\mathbf{H}_2=\mathbf{I} - 2\,\frac{(\boldsymbol{\tilde{\mu}} - \boldsymbol{\mu}_c)(\boldsymbol{\tilde{\mu}}-\boldsymbol{\mu}_c)^{\top}}{\left\|\boldsymbol{\tilde{\mu}} - \boldsymbol{\mu}_c\right\|^{2}},
\\
H(\mathbf{v}) := \mathbf{H}_2\mathbf{H}_1\mathbf{v}.
\end{gathered}
\end{equation}
Therefore, we finally compute the conditional similarity $\texttt{csim}_\text{CLAY}$ between \textit{query} image feature $\mathbf{v}_q = f_I(I_q)$ and \textit{database} image feature $\mathbf{v}_d = f_I(I_d)$ by applying $H(\cdot)$, logarithmic map $\log_{\boldsymbol{\mu}_c}({\mathbf{T}_c})$, and projection matrix $\mathbf{P}_c$ as shown in Alg.~\ref{alg:supp_method}.
Then, the final retrieved images are sorted by these values of conditional similarity.
\begin{algorithm}[t]
\caption{Conditional similarity computation}\label{alg:supp_method}
\begin{algorithmic}
\Require query feature $\mathbf{v}_q$, database image feature $\mathbf{v}_d$
\Require $\log_{\boldsymbol{\mu}_c}(\cdot)$, $\mathbf{T}_c$, $H(\cdot)$
\State $\mathbf{U}, \boldsymbol{\Sigma}, \mathbf{V}^\top \leftarrow \text{SVD}(\log_{\boldsymbol{\mu}_c}(\mathbf{T}_c))$
\State $\mathbf{P}_c \leftarrow \mathbf{V}_k \mathbf{V}_k^\top$
\State $ \mathbf{v}_{q^{\prime}} \leftarrow \mathbf{P}_c\log_{\boldsymbol{\mu}_c}(H(\mathbf{v}_q)) $
\State $ \mathbf{v}_{d^{\prime}} \leftarrow \mathbf{P}_c\log_{\boldsymbol{\mu}_c}(H(\mathbf{v}_d)) $
\State $\texttt{csim}_{\text{CLAY}} \leftarrow \dfrac{\mathbf{v}_{q^{\prime}} \cdot \mathbf{v}_{d^{\prime}}}{\|\mathbf{v}_{q^{\prime}}\| \|\mathbf{v}_{d^{\prime}}\|} $
\State \Return $ \texttt{csim}_{\text{CLAY}} $
\end{algorithmic}
\end{algorithm}

\section{CLAY-EVAL Construction Pipeline}
\label{sec:supp_clay_eval}

\begin{figure*}[t]
   \includegraphics[width=1\linewidth]{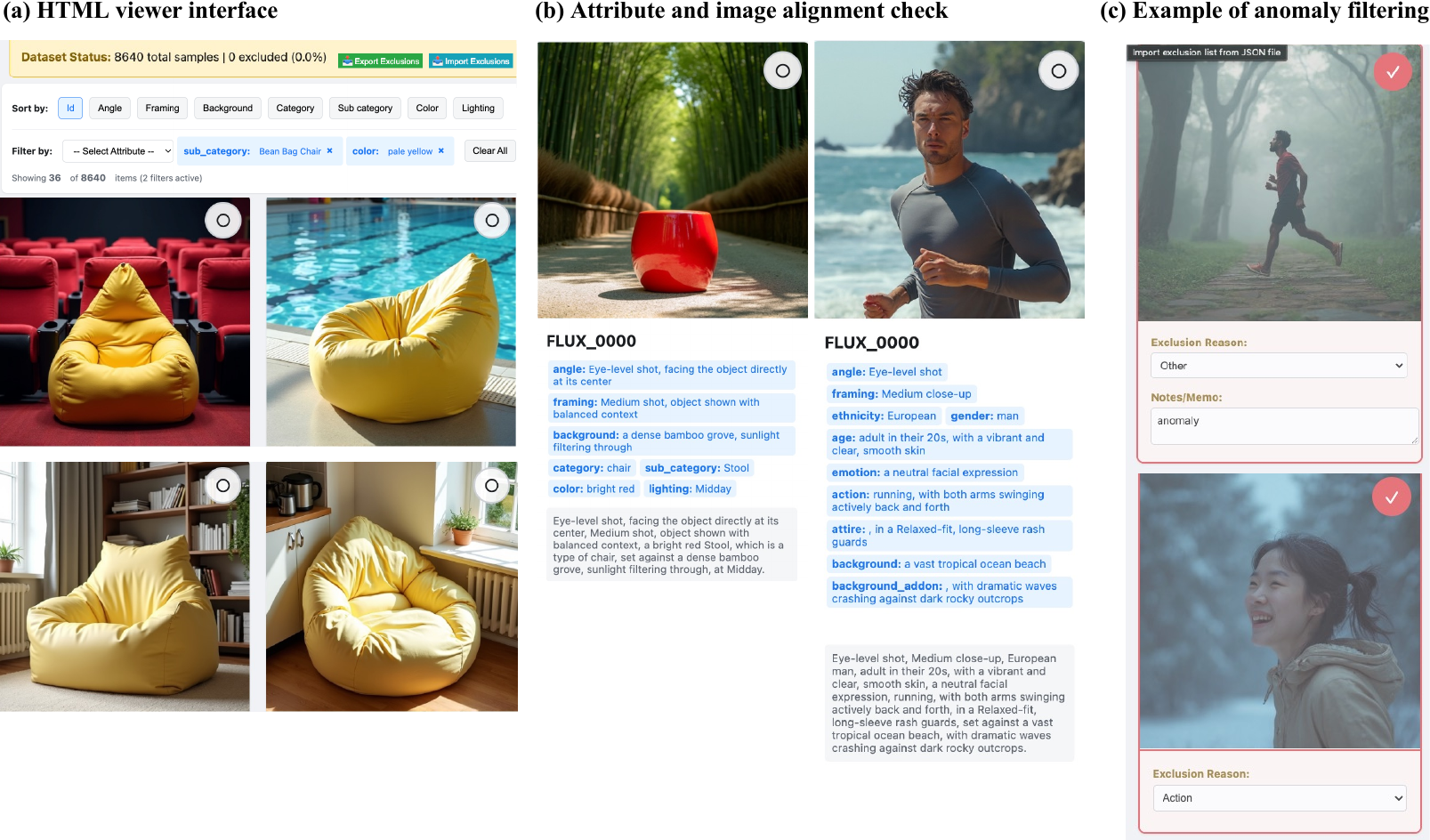}
    \caption{\textbf{HTML viewer and sample curation cases.} Illustration of the use of an HTML viewer for quality assurance and data filtering.
(a) The overall design and components of the interface used for data inspection.
The viewer example shown has been edited for visual clarity.
 Examples filtered by \texttt{sub-category}: Bean Bag Chair and \texttt{color}: pale yellow are shown.
(b) Verification that the generated image results are correctly aligned with their corresponding attributes (for both \textit{object} and \textit{human} samples) and the full text prompt used for generation.
(c) An illustration of the process for filtering out samples exhibiting anomaly or text-visual misalignment. Specific filtering cases are shown: The upper sample was excluded due to a major limb anomaly (artifact). The lower sample was excluded due to textual-visual misalignment, as the specified \texttt{action} ("running") was not discernible in the generated image.
}

\label{fig:supp_html_viewer}
\end{figure*}

In this section, we provide detailed construction pipeline of our synthetic dataset, CLAY-EVAL.
We first present overview, detail the schema of each CLAY-Object and CLAY-Human, and then present the prompts utilized for generation.

\subsection{Overview}
Our dataset construction follows a systematic four-stage process designed to ensure high-quality, diverse, and well-controlled naturalistic images.

\paragraph{Data Schema Design}
For each entity, we define core attributes (primary query conditions) and diversity attributes (used to enhance sample diversity). This stage also includes applying schema-level filtering to automatically exclude logically contradictory combinations. For example, in the CLAY-Human dataset, this step removes 288 samples for impossible scenarios (\eg, "driving" in "indoor"). Regarding the core and diversity attributes, please refer to Sec.~\ref{sec:supp_object} and \ref{sec:supp_human}.

\paragraph{Prompt Generation}
All core and diversity attributes are programmatically combined according to a structured prompt template (detailed in Sec.~\ref{sec:supp_gen_details}) to create the final, comprehensive set of text prompts.

\paragraph{Image Generation}
All samples are generated using the FLUX.1-dev model~\cite{Flux2024} with parameters specified in Sec.~\ref{sec:supp_gen_details}.

\paragraph{Curation Process}
We conduct a rigorous manual review to ensure high data quality. To facilitate this large-scale review efficiently, we utilize a custom HTML-based viewer as illustrated in Fig.~\ref{fig:supp_html_viewer}. This tool is designed to allow for selective sorting and filtering by attributes, enabling us to verify that intended compositions are visually generated correctly. It also provides functionality to select problematic samples and annotate the reasons for exclusion (managed via JSON). This step is crucial for filtering out samples that show severe text-visual misalignment, visual anomalies, or harmful content.

\subsection{Object Entity Dataset}
\label{sec:supp_object}
\paragraph{Data Schema and Construction}
For the \textit{object} entity, we select three core attributes based on their clear visual distinctiveness and hierarchical structure: \texttt{category}, \texttt{sub-category}, and \texttt{color}. To enhance visual diversity, we incorporate three additional diversity attributes: lighting, framing, and angle.

The construction pipeline begins with base combinations of core attributes. As \texttt{sub-category} (24 instances) is subordinate to \texttt{category} (4 instances), our base combinations are calculated from 10 \texttt{colors} and 24 \texttt{sub-categories}, resulting in 240 unique combinations (See Tab.~\ref{tab:core_attributes}) . To this set, we add 18 combinations of diversity attributes (3 \textit{lighting} $\times$ 3 \textit{framing} $\times$ 2 \textit{angle}; (See Tab.~\ref{tab:full_templates})), yielding 4,320 base combinations ($240 \times 18$). Finally, we sample uniformly from a pool of 71 distinct \textit{backgrounds} (detailed in Sec.~\ref{sec:supp_attr_instances}) and generate two samples for each base combination to further increase diversity, resulting in a total of 8,640 images.

\paragraph{Filtering and Final Dataset}
During our manual curation process, 1,315 samples with severe anomalies or misalignments are removed. This results in a final dataset of 7,325 samples. The final distributions for core attributes are detailed in Tab.~\ref{tab:core_attributes} (left column), while qualitative examples per attribute combination are presented in Fig.~\ref{fig:supp_clay_samples} (a).

\subsection{Human Entity Dataset}
\label{sec:supp_human}
\paragraph{Data Schema and Construction}
For the \textit{human} entity, we define \texttt{age}, \texttt{action}, and \texttt{background} as core attributes, chosen for visual identification. We also introduce five diversity attributes to ensure a balanced and representative dataset: race, gender, emotion, framing, and angle.

The base combinations consist of 3 \texttt{age} $\times$ 5 \texttt{action} $\times$ 5 \texttt{background}, total 75 combinations (See Tab.~\ref{tab:core_attributes}). We then introduce 96 diversity combinations (6 race $\times$ 2 gender $\times$ 2 emotion $\times$ 2 framing $\times$ 2 angle; (See Tab.~\ref{tab:full_templates})). This results in 7,200 total combinations ($75 \times 96$). To increase background detail without changing the total sample count, we prepare two distinct detailed descriptions ("add-ons") for each of the 5 core \texttt{backgrounds} (detailed in Sec.~\ref{sec:supp_attr_instances}) and sample uniformly between them during generation.

\begin{table*}[t]
\centering
\caption{\textbf{Prompt Templates and Core Attributes}}
\label{tab:core_attributes}
{\renewcommand{\arraystretch}{0.7}
\begin{subtable}{\textwidth}
\footnotesize
\centering
\label{tab:core_attributes_templates}
\begin{adjustbox}{width=0.95\textwidth}
\begin{tabular}{@{}l p{0.42\textwidth} | p{0.42\textwidth}@{}}
\toprule
& \multicolumn{1}{c}{\textbf{Object Entity}} & \multicolumn{1}{c}{\textbf{Human Entity}} \\
\midrule
\textbf{Template} 
    & ... \{color\} \{sub-category\}, which is a type of \{category\}, ... 
    & ... \{age\}, \{action\}, ... set against \{background\} ... \\
\midrule
\textbf{Example} 
    & ... bright red Stool, which is a type of chair, ... 
    & ... adult in their 20s, ... running, with both arms swinging actively back and forth, ... set against a vast tropical ocean beach ... \\
\midrule
\end{tabular}
\end{adjustbox}
\end{subtable}
}
\vspace{-4.5mm} % === remove vertical gap ===

\begin{subtable}{\textwidth}
\centering
\label{tab:core_attributes_attributes}
\begin{adjustbox}{width=0.95\textwidth}
\begin{tabular}{p{0.2\textwidth} p{0.3\textwidth} | p{0.2\textwidth} p{0.3\textwidth}@{}}
% \toprule
\multicolumn{4}{c}{\textbf{Core Attributes}} \\
\midrule
\textbf{Category} (4 types) 
    & Vehicle (29.19\%), Shoes (28.68\%), Bag (21.49\%), Chair (20.64\%) 
    & \textbf{Age} (3 types): &  Adult (33.08\%), Middle-aged (33.57\%), Elderly (33.36\%) \\
\midrule
\textbf{Sub-category} (24 types) 
    & Backpack (4.83\%), Bean Bag Chair (4.82\%), Duffel Bag (4.79\%), Sneakers (4.76\%), Armchair (4.64\%), High Heels (4.57\%), Tote Bag (4.57\%), Sedan (4.57\%), Loafers (4.53\%), Boots (4.52\%), Rocking Chair (4.51\%), Van (4.51\%), Motorcycle (4.45\%), Stool (4.31\%), Truck (4.20\%), Dress Shoes (4.18\%), Suitcase (4.10\%), Sandals (3.99\%), Bus (3.97\%), Bicycle (3.77\%), SUV (3.71\%), Shoulder Bag (3.19\%), Recliner (2.36\%), Slippers (2.13\%) 
    & \textbf{Action} (5 types): &  Texting (21.25\%), Reading (21.20\%), Clapping (21.08\%), Running (20.53\%), Driving (15.94\%) \\
\midrule
\textbf{Color} (10 types) : 
    & bright red (10.38\%), vivid purple (10.35\%), jet black (10.29\%), vibrant orange (10.29\%), pure white (10.28\%), pale yellow (10.21\%), deep blue (10.16\%), light pink (10.12\%), muted gray (9.30\%), dark brown (8.63\%) 
    & \textbf{Background} (5 types): & Beach (21.25\%), Snowy (20.80\%), Park (20.76\%), City (20.58\%), Indoor (16.62\%) \\
\bottomrule
\end{tabular}
\end{adjustbox}
\end{subtable}

\end{table*}

\begin{table*}[t]
\centering
\caption{\textbf{Full Prompt Templates with Diversity Attributes}}
\label{tab:full_templates}
{\renewcommand{\arraystretch}{0.7}
\begin{subtable}{\textwidth}
\footnotesize
\centering
\label{tab:full_templates_templates}
\begin{adjustbox}{width=0.95\textwidth}
\begin{tabular}{@{}l p{0.42\textwidth} | p{0.42\textwidth}@{}}
\toprule
& \multicolumn{1}{c}{\textbf{Object Entity}} 
& \multicolumn{1}{c}{\textbf{Human Entity}} \\
\midrule
\textbf{Template} 
    & "\{angle\}, \{framing\}, a \{color\} \{sub-category\}, which is a type of \{category\}, set against \{background\}, at \{lighting\}" 
    & "\{angle\}, \{framing\}, \{race\} \{gender\}, \{age\}, \{emotion\}, \{action\}, \{attire\}, set against \{background\}\{background-add-on\}" \\
\midrule
\textbf{Example} 
    & "Eye-level shot, facing the object directly at its center, Medium shot, object shown with balanced context, a bright red Stool, which is a type of chair, set against a dense bamboo grove, sunlight filtering through, at Midday."
    \newline 
    (See Fig.~\ref{fig:supp_html_viewer} (b), left column.)
    
    & "Eye-level shot, Medium close-up, European man, adult in their 20s, with a vibrant and clear, smooth skin, a neutral facial expression, running, with both arms swinging actively back and forth, in a Relaxed-fit, long-sleeve rash guards, set against a vast tropical ocean beach, with dramatic waves crashing against dark rocky outcrops." 
    \newline 
    (See Fig.~\ref{fig:supp_html_viewer} (b), right column.)
    \\
\midrule
\end{tabular}
\end{adjustbox}
\end{subtable}
}
\vspace{-4mm}

\begin{subtable}{\textwidth}
\centering
\label{tab:full_templates_diversity}
\begin{adjustbox}{width=0.95\textwidth}
\begin{tabular}{p{0.2\textwidth} p{0.3\textwidth} | p{0.2\textwidth} p{0.3\textwidth}@{}}
\multicolumn{4}{c}{\textbf{Diversity Attributes}} \\
\midrule
\textbf{Angle} (2 types) 
    & Eye-level, High-angle 
    & \textbf{Angle} (2 types) & Eye-level, Side profile\\

\textbf{Framing} (3 types) 
    & Close-up, Medium, Wide 
    & \textbf{Framing} (2 types): & Medium close-up, Full shot \\

\textbf{Background} (71 types) 
    & Urban – Global\& Iconic, & \textbf{Race} (6 types) & European, Asian, Indian, Middle Eastern, African, Latino \\
    (See Sec.~\ref{sec:supp_attr_instances}) & Indoor - Residential \& Office & \textbf{Gender} (2 types) & Man, Woman \\
    & Indoor - Public \& Commercial & \textbf{Emotion} (2 types) & Neutral, Joy \\
    & Indoor - Cultural \& Specialized & & \\
    & Indoor - Transportation & & \\
    & Nature - Specific Flora \& Weather & & \\
    & Nature - Global Biomes \& Landscapes & & \\
\textbf{Lighting} (3 types) 
    & Midday, Morning, Evening  \\

\bottomrule
\end{tabular}
\end{adjustbox}
\end{subtable}

\vspace{-2mm}
\end{table*}

\begin{figure}[t]
  \includegraphics[width=\linewidth]{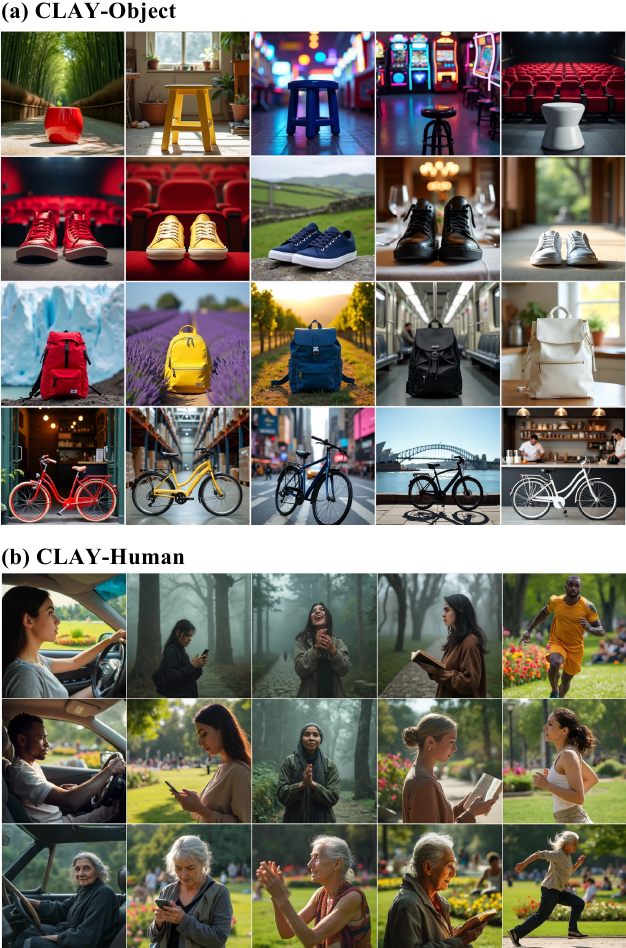}
  \caption{
    \textbf{Diverse naturalistic image samples generated according to intended attribute conditions.} We visualize image samples based on specific \textit{object} entity and \textit{human} entity attributes. (a) The grid is structured by the \texttt{Color} attribute (horizontal axis) and the \texttt{Category} attribute (vertical axis). Horizontal Axis (\texttt{Color}, 5 Columns, Left to Right): Red, Yellow, Blue, Black, White. Vertical Axis (\texttt{Category}, 4 Rows, Top to Bottom): Chair (\texttt{Sub-category}: Stool), Shoes (\texttt{Sub-category}: Sneakers), Bag (\texttt{Sub-category}: Backpack), Vehicle (\texttt{Sub-category}: Bicycle). (b) The grid is structured by the \texttt{Action} attribute (horizontal axis) and the \texttt{Age} attribute (vertical axis). Horizontal Axis (\texttt{Action}, 5 Columns, Left to Right): Driving, Texting, Clapping, Reading, Running. Vertical Axis (\texttt{Age}, 3 Rows, Top to Bottom): Adult, Middle-aged, Elderly.
  }
  \label{fig:supp_clay_samples}
  \vspace{-1.5em}
\end{figure}

\paragraph{Filtering and Final Dataset}
Following the schema-level filtering (which removed 288 logical exclusions), 6,912 combinations remain. From this set, we manually curate and remove additional 165 samples. This results in a final dataset of 6,745 samples. The final distributions for core attributes are detailed in Tab.~\ref{tab:core_attributes} (right column), and qualitative examples with each attribute combination are in Fig.~\ref{fig:supp_clay_samples} (b).

\subsection{Generation Details}
\label{sec:supp_gen_details}

\paragraph{Prompt Templates}
Based on the designed data schema, each text prompt is generated with diverse combinations of attribute instances.
% as presented in Tab.~\ref{tab:full_templates}.
% Detailed information regarding the combination of core and diversity attributes can be found in Tab.~\ref{tab:core_attributes} and Tab.~\ref{tab:full_templates}.

\paragraph{Model and Parameters}
We summarize the implementation details of FLUX below.
\begin{itemize}
    \item \textbf{Model ID:} ``black-forest-labs/FLUX.1-dev''
    \item \textbf{seed:} 42
    \item \textbf{height:} 512
    \item \textbf{width:} 512
    \item \textbf{guidance\_scale:} 0.7
    \item \textbf{num\_inference\_steps:} 50
\end{itemize}

\paragraph{Negative Prompt}
A negative prompt including the following keywords is used to suppress the generation of unintended or inappropriate images. "nude, naked, underwear, lingerie, topless, ..."

\subsection{Detailed Attribute Instances}
\label{sec:supp_attr_instances}
For visibility, several attribute instances mentioned in the paper are denoted using abbreviations (\eg, in \texttt{sub-category}: Rocking Chair $\rightarrow$ Rocking ).

\paragraph{Backgrounds for Object Entity} A pool of 71 prompts under 7 major categories is used. The type of each category and the number of contained instances are shown in Tab.~\ref{tab:full_templates}. Example from Nature - Global Biomes \& Landscapes: "rolling green hills and stone walls of the Irish countryside".

\paragraph{Background Add-ons for Human Entity} Two detailed prompts are uniformly sampled for each of the 5 base \texttt{background} types. Example from ``a vast tropical ocean beach'';  – Add-on 1: ``with palm trees swaying gently in the tropical breeze'' – Add-on 2: ``with dramatic waves crashing against dark rocky outcrops''.

\begin{table*}[t]
\caption{
\textbf{Total number of query and database images in evaluation datasets.}
}
\resizebox{\textwidth}{!}{
\begin{tabular}{lccccccccccccc}
\toprule
  & \multicolumn{1}{c}{\textbf{Action}} & \multicolumn{1}{c}{\textbf{Mood}} & \multicolumn{1}{c}{\textbf{Location}} & \multicolumn{1}{c}{\textbf{Cat}} & \multicolumn{1}{c}{\textbf{Dog}} & \multicolumn{1}{c}{\textbf{Instrument}} & \multicolumn{1}{c}{\textbf{Flower}} & \multicolumn{1}{c}{\textbf{Car}} & \multicolumn{1}{c}{\textbf{Aircraft}} & \multicolumn{1}{c}{\textbf{Food}} & \multicolumn{1}{c}{\textbf{Clevr4}} & \multicolumn{1}{c}{\textbf{CLAY-Object}} & \multicolumn{1}{c}{\textbf{CLAY-Human}} \\
\midrule
Query & 937   & 100 & 97  & 235     & 493     & 480   & 775   & 804   & 333        & 2525  & 1053  & 732    & 674   \\
Database    & 8595  & 900 & 903 & 2136    & 4485    & 4320  & 7414  & 7237  & 3000       & 22725 & 9478  & 6593   & 6071 \\
\bottomrule
\end{tabular}
}
\label{tab:supp_real_statistics}
\end{table*}

\begin{table*}[ht]
\centering
\caption{\textbf{Text instructions for comparisons across different baseline models.}} \label{tab:supp_instruction_prompts}
\resizebox{\textwidth}{!}{
\begin{tabular}{llll}
\toprule 
\textbf{Condition} & \textbf{MagicLens} & \textbf{SEARLE} & \textbf{VLM2Vec / QwenVLEmbed / InstructBLIP}\\ \midrule
\textit{Single condition} & & & \\
Cat, Dog & A \{condition\} of the same breed & a photo of the same \{condition\} species of \$ & What breed of \{condition\} is in the image? \\
Car, Aircraft & The same \{condition\} model & a photo of the same \{condition\} model of \$ & What model of \{condition\} is in the image? \\
Count & The same object \{condition\} & a photo of the same object \{condition\} of \$ & How many objects are in the image? \\
Age & The same human \{condition\} & a photo of the same human \{condition\} of \$ & What age of human in the image? \\
Color, Category (Object) & The same object \{condition\} & a photo of the same object \{condition\} of \$ & What type of object \{condition\} is in the image? \\
Etc. & The same (type of) \{condition\} & a photo of the same (type of) \{condition\} of \$ & What type of \{condition\} is in the image? \\ \midrule
\textit{Multi condition} & & & \\ 
Age, Action & - & - & What age of human and what type of action is in the image? \\ 
Age, Background & - & - & What age of human and what type of background is in the image? \\ 
Age, Action, Background & - & - & What age of human and what type of action, background is in the image? \\ 
Etc. & - & - & What type of \{condition1\} and \{condition2\} is in the image? \\ 
\bottomrule
\end{tabular}
}
\end{table*}

\section{Experimental Details}
\label{sec:supp_exp_details}

\subsection{Experimental Setup}
For evaluation, we consider the real-world fine-grained image classification datasets for single conditional retrieval setting, including Stanford40~\cite{yao2011human}, OxfordPets~\cite{parkhi12a}, PPMI~\cite{yao2010grouplet, wang2010locality}, Flowers102~\cite{Nilsback08}, StanfordCars~\cite{Krause_2013_stanfordcars}, FGVC-Aircraft~\cite{maji2013aircraft}, and Food-101~\cite{bossard2014food}.
We randomly split each dataset into a 1:9 ratio for the construction of \textit{query} and \textit{database}, the final number of images used to evaluation are presented in Tab.~\ref{tab:supp_real_statistics}.

\subsection{Implementation Details}
We request Large Language Model to generate $n=100$ prompts, and encode them into the text encoder of Vision Language Model.
However, we find that the outputs do not strictly follow the specified total counts. Nevertheless, we utilize the entire set of generated prompts in our experiments.
In the process of Singular Value Decomposition, we truncate the top-$k$ singular vectors to obtain projection matrix following previous works~\cite{dorfman2025ip, nguyen2025csd}.
In our experiments, we set $k=50$.
For our comparison methods~\cite{zhang2024magiclens,baldrati2023zero,dai2023instructblip,qwen3vlembedding,meng2025vlm2vecv2} we set the instructions for describing the text condition as shown in Tab.~\ref{tab:supp_instruction_prompts}.
A few instruction examples in single condition setup are selected by following FocalLens~\cite{hsieh2025focallens}, and the remaining samples are chosen manually.
In addition, for GeneCIS~\cite{vaze2023genecis}, we set the condition texts as those in the \textit{condition} column of Tab.~\ref{tab:supp_instruction_prompts}.

\begin{table*}[t]
\caption{
\textbf{Quantitative comparison of mean Average Precision~(mAP) across VLMs with ViT-L.}
}
\centering
\begin{subtable}{0.99\textwidth}
\resizebox{\textwidth}{!}{
\begin{tabular}{lcccccccccc}
\toprule
\multicolumn{1}{c}{} & \multicolumn{3}{c}{\textbf{Stanford40}} & \multicolumn{7}{c}{\textbf{Fine-grained Image Classification}}     \\ 
\cmidrule(lr){2-4} \cmidrule(lr){5-11}
Method & Action & Location & Mood & Cat & Dog & Instrument & Flower & Car & Aircraft & Food \\
\midrule
CLIP-L & 45.3 & 44.1 & 54.0 & 46.7 & 49.1 & 35.9 & 83.6 & 42.4 & 29.0 & 59.0 \\
SigLIP-L& 56.5 & 51.3 & 56.7 & 63.1 & 68.8 & 53.0 & 88.9 & 76.5 & 60.2 & 69.7 \\
Ours (CLIP-L) & 68.9 & 55.5 & \textbf{63.2} & 80.1 & 87.0 & 70.8 & 89.7 & 65.5 & 43.4 & 71.5 \\
Ours (SigLIP-L) & \textbf{72.2} & \textbf{60.0} & 62.0 & \textbf{87.8} & \textbf{89.8} & \textbf{79.5} & \textbf{97.7} & \textbf{86.3} & \textbf{72.7} & \textbf{76.6} \\
\bottomrule
\end{tabular}
}
\captionsetup{font=small} \caption{{real-world datasets}.} 
\end{subtable}

\vspace{6pt}

\begin{subtable}{0.99\textwidth}
\resizebox{\textwidth}{!}{
\begin{tabular}{lcccccccccc}
\toprule
& \multicolumn{4}{c}{\textbf{Clevr4}} & \multicolumn{3}{c}{\textbf{CLAY-Object}} & \multicolumn{3}{c}{\textbf{CLAY-Human}} \\
\cmidrule(lr){2-5} \cmidrule(lr){6-8} \cmidrule(lr){9-11}
Method & Shape & Color & Texture & Count & Color & Category & Subcategory & Age & Action & Background \\
\midrule
CLIP-L & 71.1 & 17.1 & 18.7 & 13.5 & 12.9 & 70.3 & 48.8 & 47.3 & 55.0 & 39.1 \\
SigLIP-L & 83.6 & 18.7 & 19.4 & 13.0 & 15.3 & 67.5 & 60.8 & 48.1 & 57.6 & 54.7 \\
Ours (CLIP-L) & 79.5 & 64.8 & 24.9 & 19.9 & 47.2 & \textbf{95.1} & 78.2 & \textbf{68.9} & \textbf{81.9} & 82.3 \\
Ours (SigLIP-L) & \textbf{93.1} &\textbf{67.6} & \textbf{25.6} & \textbf{25.8} & \textbf{64.9} & 94.0 & \textbf{83.3} & 67.6 & 81.42 & \textbf{91.6} \\
\bottomrule
\end{tabular}
}
\captionsetup{font=small} \caption{{synthetic datasets}.} 
\end{subtable}
\label{tab:supp_backbone}
\end{table*}

\section{Additional Results}
\label{sec:supp_additional_results}
\paragraph{Backbone variation in VLMs}
To demonstrate the effectiveness of our method under large models with ViT-L, we evaluate retrieval performance varying the backbone model on single conditional datasets.
Table~\ref{tab:supp_backbone} demonstrates that our method shows substantial gains in mAP compared to the baseline.
This suggests that our method has the potential to provide greater improvements when applied to large models.

\begin{table}[h]
\caption{\textbf{Quantitative comparison of Recall@1,2,3 on GeneCIS benchmark.} 
We evaluate the retrieval performance on the ``Focus Attribute'' subset, where each query has a single ground-truth match. 
Recall@1, @2, and @3 are reported as the performance metric. For CIR methods, we report the values provided from MagicLens~\cite{zhang2024magiclens}.
}
\vspace{-0.5em}
\centering
\footnotesize
\renewcommand{\arraystretch}{1.0} 
    \resizebox{1\linewidth}{!}{
    \begin{tabular}{m{2cm} ccc}
    \toprule
    Method  &  Recall@1 &Recall@2 & Recall@3\\
    \midrule
    CLIP-B  &  17.8 & 30.0 & 40.4 \\ \midrule
    GeneCIS  & 19.5 & 31.8 & 42.2 \\
    CIReVL & 17.9 & 29.4 & 40.4 \\
    MagicLens  & 15.5 & 28.4 & 39.1 \\
    SEARLE  & 17.0 & 29.7 & 40.7 \\ \midrule
    Ours~(CLIP-B)   &\textbf{24.4}& \textbf{38.3} &\textbf{50.5} \\
    \bottomrule
    \end{tabular}
    }
\label{tab:genecis}
\end{table}
\paragraph{Comparison on GeneCIS benchmark}
We also compare our method with Composed Image Retrieval methods~\cite{zhang2024magiclens, karthik2024vision, baldrati2023zero} on the GeneCIS benchmark ``Focus Attribute'' subset.
We report Recall at $k$ following previous work~\cite{vaze2023genecis}.
Table~\ref{tab:genecis} shows CLAY still achieves the best performance when utilizing the same or even smaller backbone (\ie, ViT-B). 

\begin{table}[t]
\centering
\caption{\textbf{Quantitative result of multi-conditioned retrieval}} \label{screeplot}
  \vspace{0pt} 
  \centering
  \label{tab:supp_multicondreal}
  \resizebox{\linewidth}{!}{%
    \begin{tabular}{lcc}
      \toprule
      & \multicolumn{2}{c}{\textbf{Stanford40}} \\
      & \makecell{\textbf{Action}~+~\textbf{Mood}} &
        \makecell{\textbf{Action}~+~\textbf{Location}} \\
      \midrule
      GeneCIS & 54.1 & 47.2 \\
      GeneCIS$^\mathbf{\dagger}$ & 65.2 & 55.5 \\
      Ours (CLIP-B) & 68.3 & 57.9 \\
      Ours (CLIP-L) & \textbf{72.0} & \textbf{58.5} \\
      \bottomrule
    \end{tabular}%
  }
\end{table}

\paragraph{Multi-conditional retrieval on real dataset}
We report multi-conditioned retrieval results of GeneCIS and our method on Stanford40 dataset, using samples labeled with multiple attribute pairs. 
While GeneCIS cannot input multi-conditioned inputs due to its training setup, we nevertheless applied such inputs for experimental evaluation.
In Tab.~\ref{tab:supp_multicondreal}, we find that CLAY achieves higher retrieval performance than GeneCIS, highlighting the effectiveness of our conditioning process on multiple inputs.

\begin{figure}[t]
  \vspace{0pt} 
  \centering
  \includegraphics[width=0.8\linewidth]{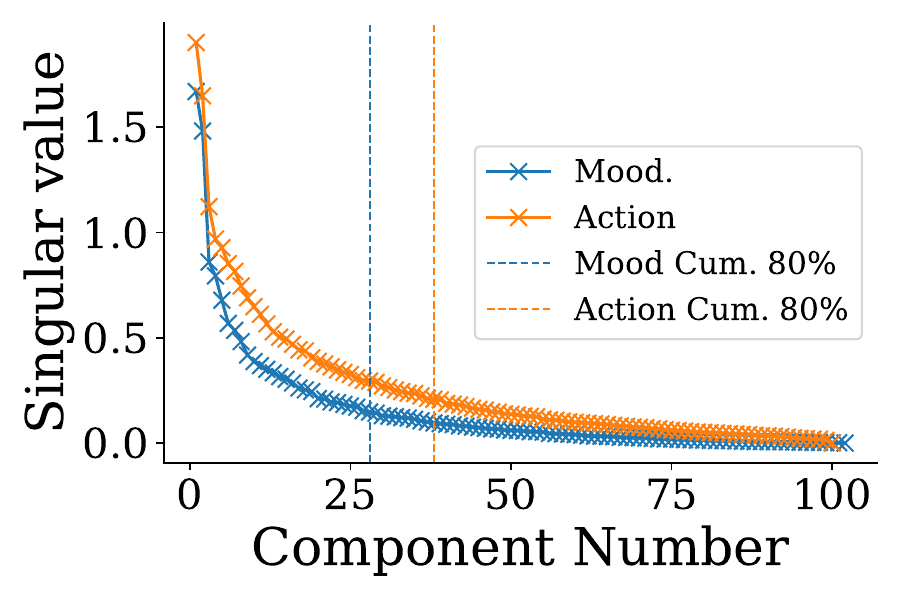}
  \vspace{-1em} 
  \caption{\textbf{Scree Plot}} \label{fig:supp_scree}
  \vspace{-1.4em} 
\end{figure}

\paragraph{Scree plot of singular values}
We provide scree plot of singular values in Fig.~\ref{fig:supp_scree}. The top-50 singular values explain over 80\% of the variance, implying that the condition-related features can be approximated by our design.

\begin{figure*}[t]
   \includegraphics[width=0.95\linewidth]{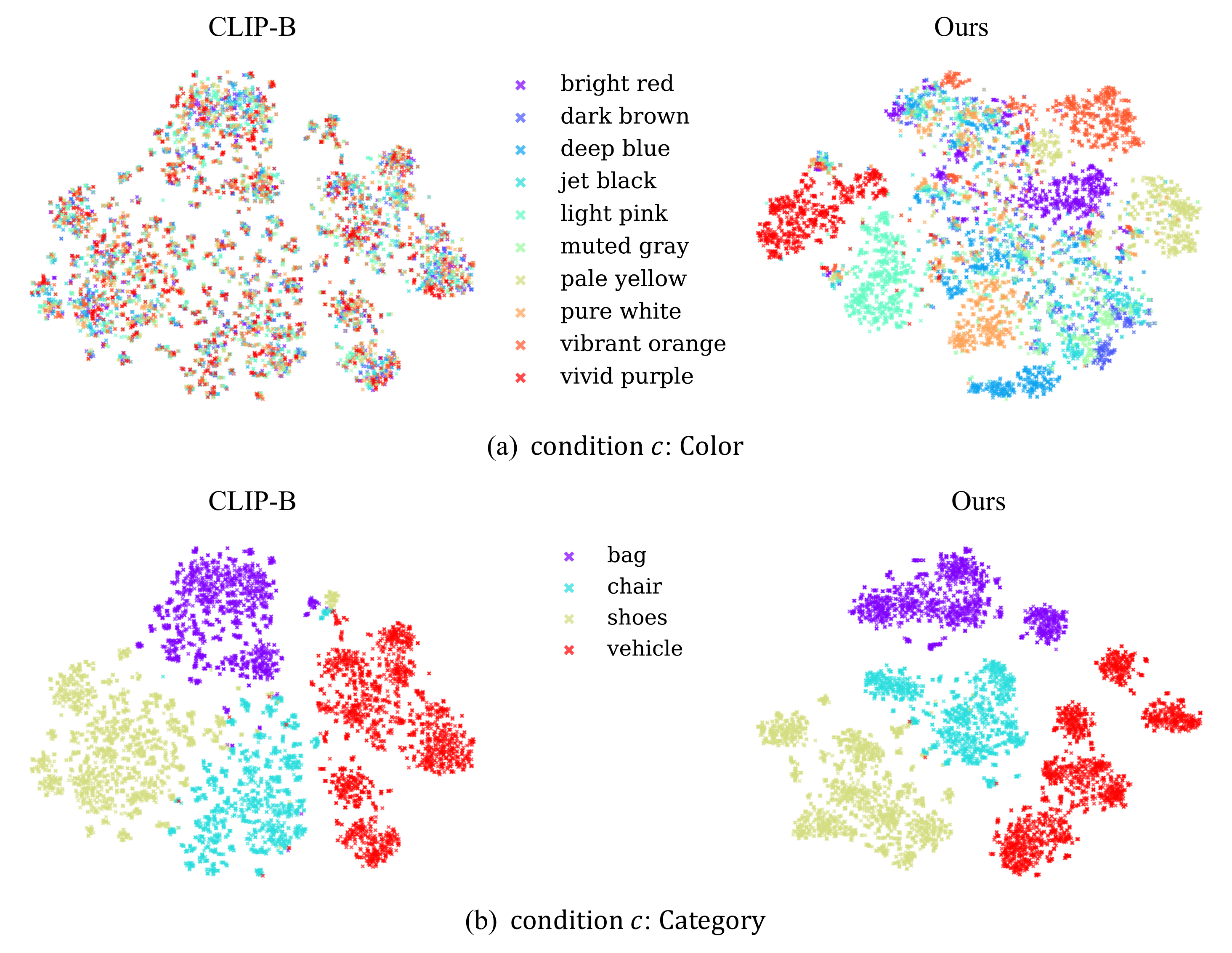}
\caption{\textbf{T-SNE visualization on CLAY-EVAL-Object dataset.}}
\label{fig:supp_tsne_synthobject}
\end{figure*}

\begin{figure*}[t]
   \includegraphics[width=0.95\linewidth]{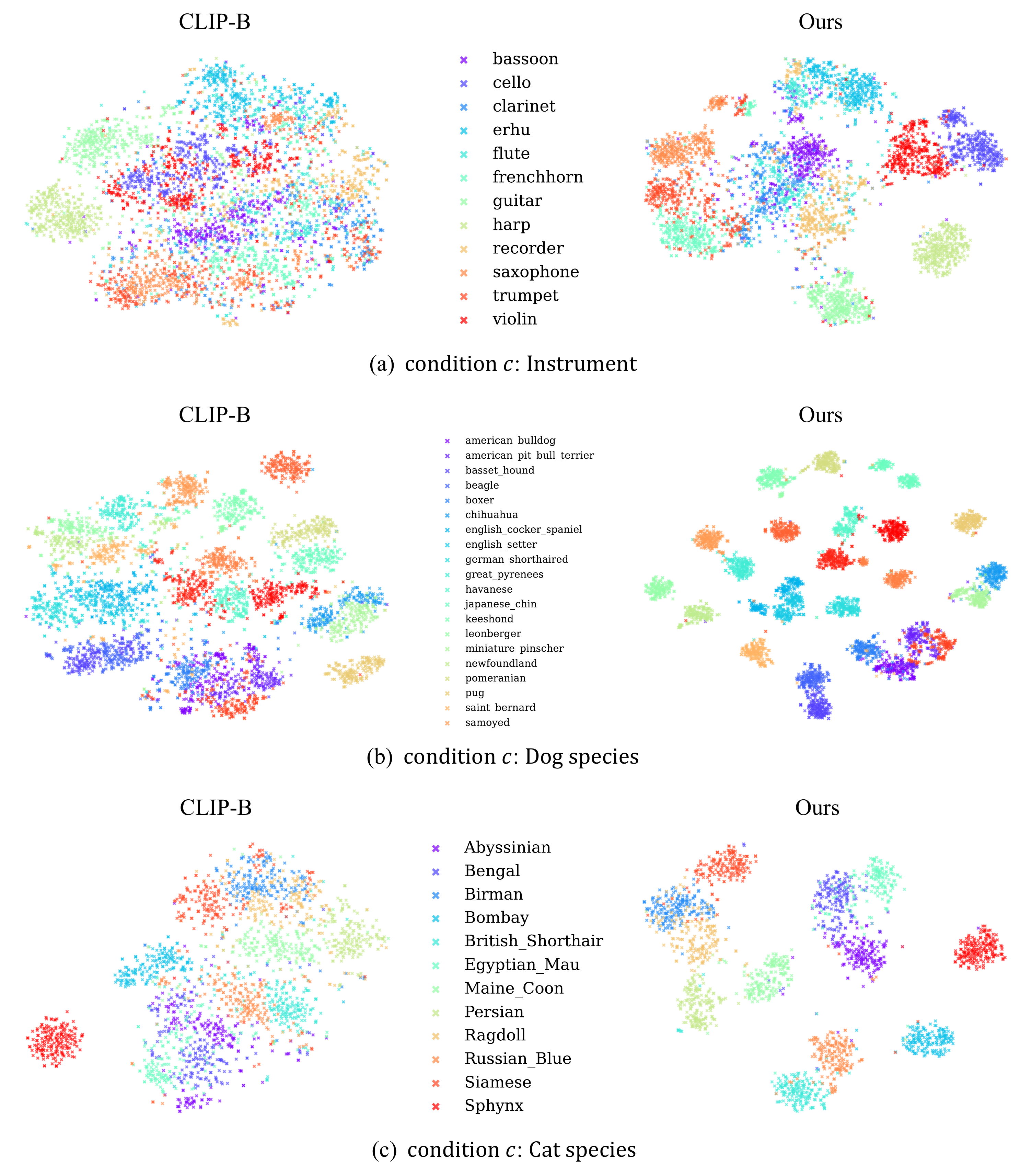}
\caption{\textbf{T-SNE visualization on fine-grained classification datasets.}}
\label{fig:supp_tsne_classfic}
\end{figure*}

\begin{figure*}[t]
   \includegraphics[width=0.95\linewidth]{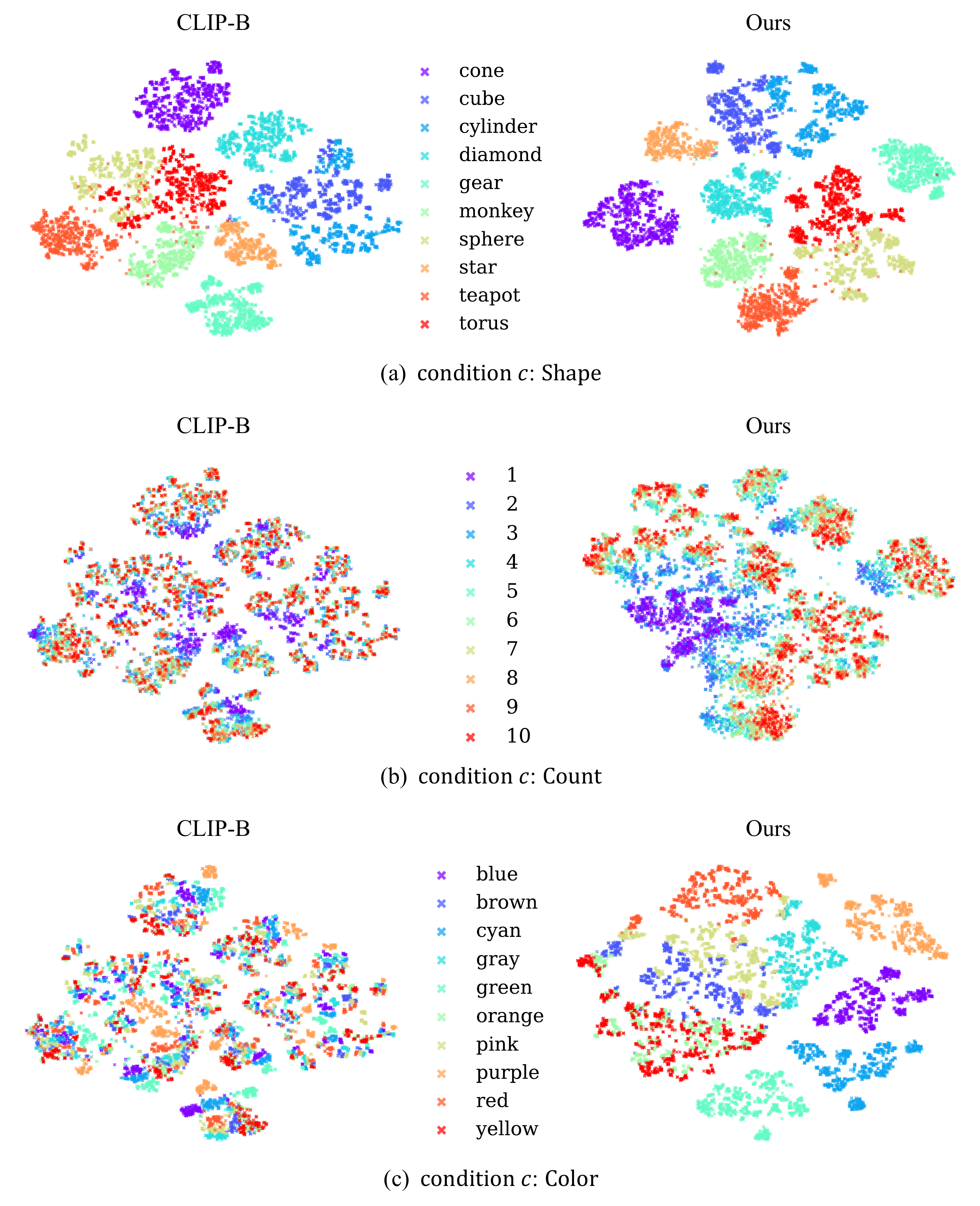}
\caption{\textbf{T-SNE visualization on Clevr4 dataset.}}
\label{fig:supp_tsne_clevr4}
\vspace{-1.5em}
\end{figure*}

\paragraph{Additional t-SNE plots}
We visualize additional t-SNE results of base model~(CLIP-B) and ours~(CLIP-B) with \textit{database} features in evaluation datasets.
Figure~\ref{fig:supp_tsne_synthobject} shows the representations under varying conditions in CLAY-EVAL-Object dataset, highlighting the effectiveness of our method for separating the conditional visual features.
Furthermore, in Fig.~\ref{fig:supp_tsne_classfic} and \ref{fig:supp_tsne_clevr4}, our method can adaptively generate well-separated and structured clusters compared to the base model.
Meanwhile, Fig~\ref{fig:supp_tsne_clevr4}~(b) shows the representations under \texttt{count} condition in Clevr4~\cite{johnson2017clevr}.
As shown in this figure, the image features corresponding to numbers 1 to 10 are aligned in a radial direction from the bottom left.
This indicates that our method can enhance the rankable property~\cite{sonthalia2025rankability} in the original model representation space.

\begin{figure*}[t]
   \includegraphics[width=1\linewidth]{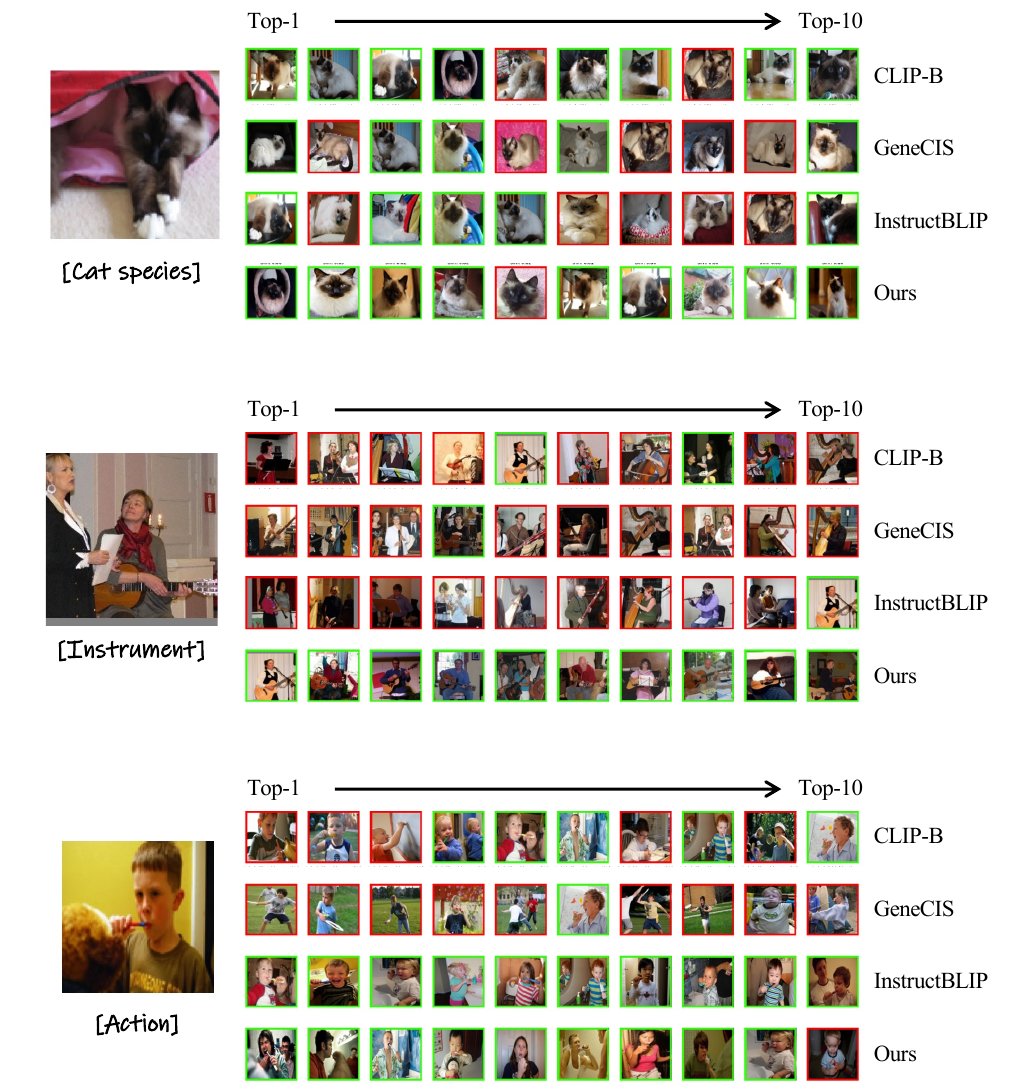}
\caption{\textbf{Qualitative results on real datasets.}}
\label{fig:supp_qual1}
\end{figure*}

\begin{figure*}[t]
   \includegraphics[width=1\linewidth]{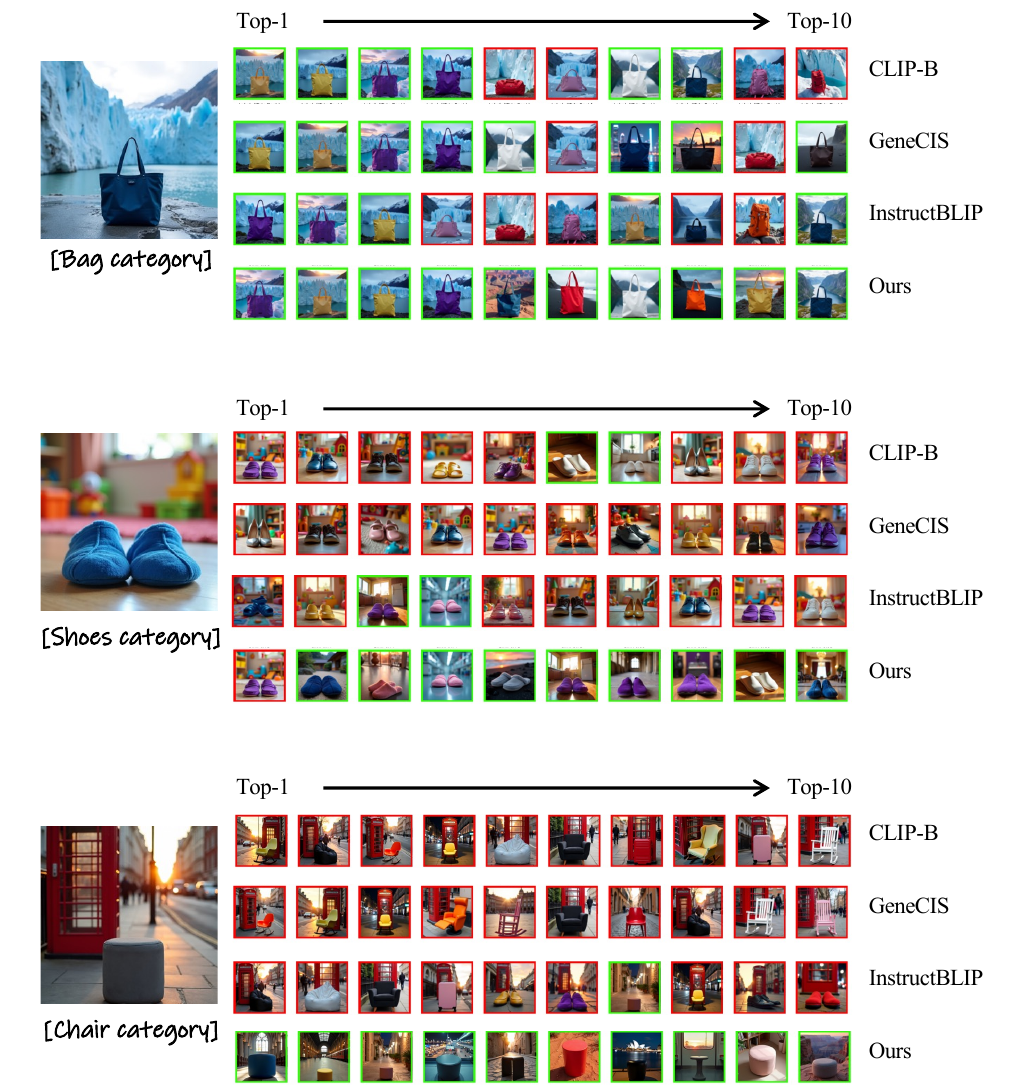}
\caption{\textbf{Qualitative results on synthetic datasets.}}
\label{fig:supp_qual2}
\end{figure*}

\paragraph{Further qualitative results}
In Fig.~\ref{fig:supp_qual1} and Fig.~\ref{fig:supp_qual2}, we show additional qualitative results of our method and comparisons for diverse conditions.
Given a \textit{query} image and a specific condition, our method reflects the condition and retrieves \textit{database} images accordingly, even distinguishing the differences within the same category~(such as \texttt{bag category, shoes category}).
Moreover, under the \texttt{action} condition, our method can disentangle the action attribute with the attributes of the subject~(\eg, age, gender), whereas the baseline struggles to retrieve diverse images.
These results demonstrate the condition fidelity of our approach under a wide range of conditions.

\section{Limitations}
\label{sec:supp_limitations}
While our method achieves substantial gains in retrieval performance, we have a few limitations.
First, our work focuses on the core attributes present in the image, which may limit its applicability to the general scenarios, especially when the task involves focusing on multiple objects.
Secondly, as our method depends on the feed-forward visual features extracted from VLM, it may not recover attributes that are not encoded in the original representations.
For example, prior work~\cite{abbasi2025analyzing} shows that the CLIP image encoder favors larger objects, suggesting that the small objects may not be fully captured and thus our method may fail in this case.
However, these limitations can be alleviated by allowing users to specify their conditions in a more direct manner~(\eg, segmentation masks), which may help recover the attributes and enable our method to perform retrieval effectively.

\section{Ethical Considerations}
Our synthetic dataset CLAY-EVAL involves human images fully generated by image generation model, FLUX~\cite{Flux2024}.
It does not include any real individuals, and we designed both text prompts and generation process to ensure this.
The demographic categories referenced—specifically European, Asian, Indian, Middle Eastern, African, and Latino (Hispanic)—are sourced from Fair Diffusion~\cite{friedrich2023fair} and FairFace~\cite{karkkainen2021fairface}.
We acknowledge the limitations of these categories, which represent broad, oversimplified groupings and contain inconsistent levels of granularity (e.g., mixing continental, national, and ethnic labels).
These labels were used solely for the purpose of ensuring sample diversity in this paper.
Our model is not supplied with and does not make use of this information at any stage of training or inference.
To promote safe and responsible use, we performed manual screening step to remove samples containing harmful content.
Nevertheless, we recognize the misuse of synthetic human images remains, and we encourage users to utilize the dataset with responsibility and follow the ethical guidelines.

\clearpage

\clearpage

\clearpage

\clearpage

\end{document}